\documentclass[journal,onecolumn,draftclsnofoot,web]{ieeecolor}
\usepackage{tmi}
\usepackage[utf8]{inputenc}
\usepackage{subfig}
\usepackage{graphicx}
\usepackage{tabularx}
\usepackage{multirow}
\usepackage{booktabs}
\usepackage{amsmath,amssymb,amsfonts}
\usepackage{bm}
\usepackage{makecell}
\usepackage[normalem]{ulem}
\usepackage[font=small,labelfont=bf]{caption}
\usepackage{cite}

\newcolumntype{L}[1]{>{\raggedright\let\newline\\\arraybackslash\hspace{0pt}}m{#1}}
\newcolumntype{C}[1]{>{\centering\let\newline\\\arraybackslash\hspace{0pt}}m{#1}}
\newcolumntype{R}[1]{>{\raggedleft\let\newline\\\arraybackslash\hspace{0pt}}m{#1}}

\graphicspath{{figures/}}

\def\argmin{\mathop{\mathrm{argmin}}}

\def\vec#1{\ensuremath{\bm{\mathit{#1}}}}

\begin{document}

\title{Self-supervised Learning from 100 Million Medical Images}

\author{Florin~C.~Ghesu, \IEEEmembership{Member, IEEE}, Bogdan~Georgescu, Awais~Mansoor, Youngjin~Yoo,\\ Dominik~Neumann, Pragneshkumar~Patel, R.S.~Vishwanath, James~M.~Balter, Yue~Cao,\\Sasa~Grbic, and Dorin~Comaniciu, \IEEEmembership{Fellow, IEEE}

\thanks{Florin C. Ghesu, Bogdan Georgescu, Awais Mansoor, Youngjin Yoo, Pragneshkumar Patel, Sasa Grbic and Dorin Comaniciu are with Siemens Healthineers, Digital Technology and Innovation, Princeton, NJ, USA. (email:florin.ghesu@siemens-healthineers.com)}
\thanks{Dominik Neumann is with Siemens Healthineers, Digital Technology and Innovation, Erlangen, Germany}
\thanks{R.S. Vishwanath is with Siemens Healthineers, Digital Technology and Innovation, Bangalore, India}
\thanks{James M. Balter and Yue Cao are with the University of Michigan, Department of Radiation Oncology, Ann Arbor, MI, USA}}

\maketitle

\begin{abstract}
Building accurate and robust artificial intelligence systems for medical image assessment requires not only the research and design of advanced deep learning models but also the creation of large and curated sets of annotated training examples. Constructing such datasets, however, is often very costly -- due to the complex nature of annotation tasks and the high level of expertise required for the interpretation of medical images (e.g., expert radiologists). To counter this limitation, we propose a method for self-supervised learning of rich image features based on contrastive learning and online feature clustering. For this purpose we leverage large training datasets of over 100,000,000 medical images of various modalities, including radiography, computed tomography (CT), magnetic resonance (MR) imaging and ultrasonography. We propose to use these features to guide model training in supervised and hybrid self-supervised/supervised regime on various downstream tasks. We highlight a number of advantages of this strategy on challenging image assessment problems in radiography, CT and MR: 1) Significant increase in accuracy compared to the state-of-the-art (e.g., AUC boost of 3-7\% for detection of abnormalities from chest radiography scans and hemorrhage detection on brain CT); 2) Acceleration of model convergence during training by up to 85\% compared to using no pretraining (e.g., 83\% when training a model for detection of brain metastases in MR scans); 3) Increase in robustness to various image augmentations, such as intensity variations, rotations or scaling reflective of data variation seen in the field.
\end{abstract}

\begin{IEEEkeywords}
Self-supervised learning, clustering, semi-supervised learning, abnormality assessment
\end{IEEEkeywords}

\section{Introduction}
\IEEEPARstart{S}{elf}-supervised learning has enjoyed much attention in recent years in the vision research community, with methods powered by large amounts of data nearing the accuracy level of state-of-the-art supervised learning strategies on well known benchmarks such as ImageNet~\cite{Caron2020, Chen2020, ImageNet}. Moreover, they demonstrate that one can use visual representations derived through self-supervised learning to guide regular downstream supervised learning and achieve increased performance (e.g., via transfer learning).

Only few studies have investigated the impact of self-supervised learning in the medical image analysis domain (e.g.,~\cite{Zhou2019, Chaitanya2020, Nguyen2020}) -- a field where the development of AI technologies is impacted by a high cost of annotations (often requiring expert radiologists precision) and scarcity of medical imaging data. These solutions are generally limited in their design to focus on architectures for segmentation (i.e., encoder-decoder) and do not support deep architectures often used for classification or detection~\cite{He2016, Tian2019}. In addition, these methods do not exploit truly large datasets and are at best trained with thousands or hundreds of thousands of cases - the same range as many systems trained with supervised learning~\cite{Guendel2021}. In this work we overcome these limitations by proposing a method for self-supervised learning from medical image data which enables the training of classification-optimized architectures. In particular, we make a first step towards truly big-data training and break the barrier of 100,000,000 training images.
%

The contributions of the paper are the following:
\begin{itemize}
    \item We propose a method for self-supervised learning based on contrastive learning~\cite{Hadsell2006} and online feature clustering~\cite{Caron2020}. The method enables hybrid self-supervised/ supervised learning from multi-modality data, and is applicable to 2D and 3D image data. As core part of the system, we propose a new set of image transformation operations optimized for medical image data. Closest to our work is the contribution of Caron et al.~\cite{Caron2020}.
    \item We conduct large scale self-supervised training experiments, including a dataset of over 1,300,000 X-rays and a dataset of over 105,000,000 multi-modality image data (including X-ray, CT, MR, US). To the best of our knowledge this represents the largest machine learning experiment to date focused on medical image data that has been reported in the literature.
    \item We perform a rigorous validation of the method on three medical computer aided diagnosis (CAD) problems: 1) Chest radiography abnormality assessment; 2) Brain metastasis detection in MR; and 3) Brain hemorrhage detection in CT data. For this purpose, we use challenging test datasets that are reflective of real clinical practice and with highly curated annotations derived by consensus of multiple expert radiologists. This is an essential step in obtaining an accurate assessment of performance. We intentionally avoid public datasets such as ChestX-ray8~\cite{NIHdata} with reported suboptimal image quality and label error rates of 65 - 85\% in terms of sensitivity~\cite{Guendel2021}.
    \item We demonstrate that by using the proposed method one can achieve a considerable performance increase on all the previously enumerated tasks, i.e., significant accuracy increase (average of 6-8\% AUC), robustness gain, and acceleration of model training convergence (up to 85\%).
\end{itemize}

The paper is organized as follows: Section~\ref{sec:related} provides an overview of related work, with the last subsection focusing on recent developments for self-supervised learning in the medical imaging domain; Section~\ref{sec:method} describes the proposed method followed by Section~\ref{sec:experiments} in which we present the experiments on various abnormality detection problems based on different 2D/3D image modalities. Finally, Section~\ref{sec:conclusion} concludes the paper with a summary and outlook on future work.

\section{Background and Motivation}
\label{sec:related}

\subsection{Self-Supervised Learning by Contrastive Learning}

Proposed as a principled approach for dimensionality reduction~\cite{Hadsell2006}, contrastive learning based on invariant input transformations has become a key optimization strategy for self-supervised feature learning. Using various transformations of the input data which determine a series of surrogate classes, Dosovitskiy et al.~\cite{Dosovitskiy2014} propose a supervised discriminative learning approach as a means to learn robust features from unlabeled data. In contrast, Bojanowski et al.~\cite{Bojanowski2017} learn a supervised mapping to a set of target deep representations sampled from an uninformative distribution, referred to as \emph{noise-as-targets}. Using this strategy, they argue that one can avoid learning trivial feature sets or the effects of feature collapse. One limitation of instance learning discussed also in~\cite{Dosovitskiy2014} is the intractable number of classes which is proportional to the number of instances. Wu et al.~\cite{Wu2018} address this limitation using a non-parametric approach which constructs a memory bank to store the target instance representations and applies noise-contrastive estimation (NCE)~\cite{Gutmann2010} to compare instances. A memory bank is used also by Zhuang et al.~\cite{Zhuang2019} for their local aggregation scheme, designed to optimize the instance representation such the similar data samples are clustered, while dissimilar ones become separated in the target manifold. Recently, Kaiming et al.~\cite{Kaiming2020} proposed to replace the memory bank with a momentum encoder coupled with a queue to generate and store representations for contrastive learning. In contrast, Hjelm et al.~\cite{Hjelm2019} propose to use mutual information maximization based on NCE~\cite{Gutmann2010} for unsupervised feature learning - applying adversarial learning to constrain the representation according to a given prior. Bachman et al.~\cite{Bachman2019} extend the approach to optimize the mutual information on multiple feature scales based on so called multiple views, i.e., different augmentations of the input. Tian et al.~\cite{Tian2020} further extend the method proposed by Hjelm et al.~\cite{Hjelm2019} to support more than two views for an improved performance. Similar principles are applied by Henaff et al.~\cite{Henaff2020} using contrastive predictive coding to learn deep representations from a spatial decomposition of the input image.

\subsection{Self-Supervised Learning by Clustering}

Unsupervised representation learning using clustering~\cite{Bautista2016, Caron2018, Zhuang2019, Caron2020} is a common alternative to instance learning and contrastive learning. Caron et al.~\cite{Caron2018} propose DeepCluster, an end-to-end unsupervised feature learning approach using the k-means clustering algorithm as optimization criteria. Coupled with the self-supervised learning method presented in~\cite{Gidaris2018}, the method is further enhanced in~\cite{Caron2019} to effectively scale to large uncurated datasets. In their approach Xueting et al.~\cite{Xueting2020} also rely on the k-means algorithm, but in a two stage approach: first cluster assignments are computed from a pretrained model and used as pseudo-labels in the second stage for feature learning. In contrast, Huang et al.~\cite{Huang2019} introduce anchor neighborhood discovery - a divide-and-conquer strategy coupled with curriculum learning for effective sample clustering. Using this optimization criteria they demonstrate that one can learn representative deep features in an end-to-end manner. Different from this, Asano et al.~\cite{Asano2020} propose an effective algorithm for simultaneous feature learning and label inference by maximizing the mutual information between data indices and labels in the form of an optimal transport problem.

\subsection{Learning from Pretext Task}

Another formulation for self-supervised learning reduces the problem to learning from a supervised signal that is artificially constructed to model a pretext task, e.g., solving a Jigsaw puzzle~\cite{Noroozi2016, Kim2018}. Agrawal et al.~\cite{Agrawal2015} propose to use egomotion as supervision, demonstrating that the features learnt from movement prediction are superior to features learned from traditional image labels. Similarly, Misra et al.~\cite{Misra2016} learn feature representations by estimating the correct temporal order of frames in video captures. Inspired by early approaches for landmark detection via coordinate regression~\cite{Zhou2007}, Doersch et al.~\cite{Doersch2015} propose to use visual context as supervised signal, learning to estimate the relative position of pairs of patches extracted from given unlabeled images. Noroozi et al.~\cite{Noroozi2016} propose as pretext task and artificial Jigsaw puzzle of image tiles. They demonstrate that one can train a deep learning model (in the form of a context free network) to solve the puzzle and thereby learn rich semantic features. An alternative strategy is feature learning by inpainting using context encoders trained with an adversarial optimization criterion~\cite{Pathak2016}. Finally, Larsson et al.~\cite{Larsson2016} use colorization as pretext task, learning to estimate a per-pixel color histogram.

\subsection{Self-Supervised Learning in the Medical Domain}

Similar principles for self-supervised feature learning are applied in medical image analysis to improve the accuracy and robustness of downstream tasks, e.g., abnormality classification or anatomy segmentation~\cite{Zhou2019, Navarro2021}. For instance, Chen et al.~\cite{Chen2019} propose a commonly known restoration strategy for feature learning from images with artificially swapped local patches. In contrast, Zhou et al.~\cite{Zhou2019} apply various image manipulation steps (nonlinear intensity transformation, local pixel shuffling or in/out-painting) and train an encoder-decoder architecture to reconstruct the original image information thereby learning rich semantic features in an unsupervised way. With focus on volumetric anatomy segmentation, Chaitanya et al.~\cite{Chaitanya2020} introduce a framework for self-supervised learning based on a hybrid contrastive loss, that learns both global and local image representations. For the same application, Nguyen et al.~\cite{Nguyen2020} propose to use spatial awareness as signal for self-supervised learning -- learning to predict the displacement of different image slices after random swaps of image patches between slices. Finally, Azizi et al.~\cite{Azizi2021} rely on the contrastive learning based method proposed in~\cite{Chen2020} to pretrain features and improve the accuracy of various downstream classification tasks from radiography or dermatology images.

\section{Proposed Method}
\label{sec:method}

\begin{figure}[t]
\includegraphics[width=8.8cm]{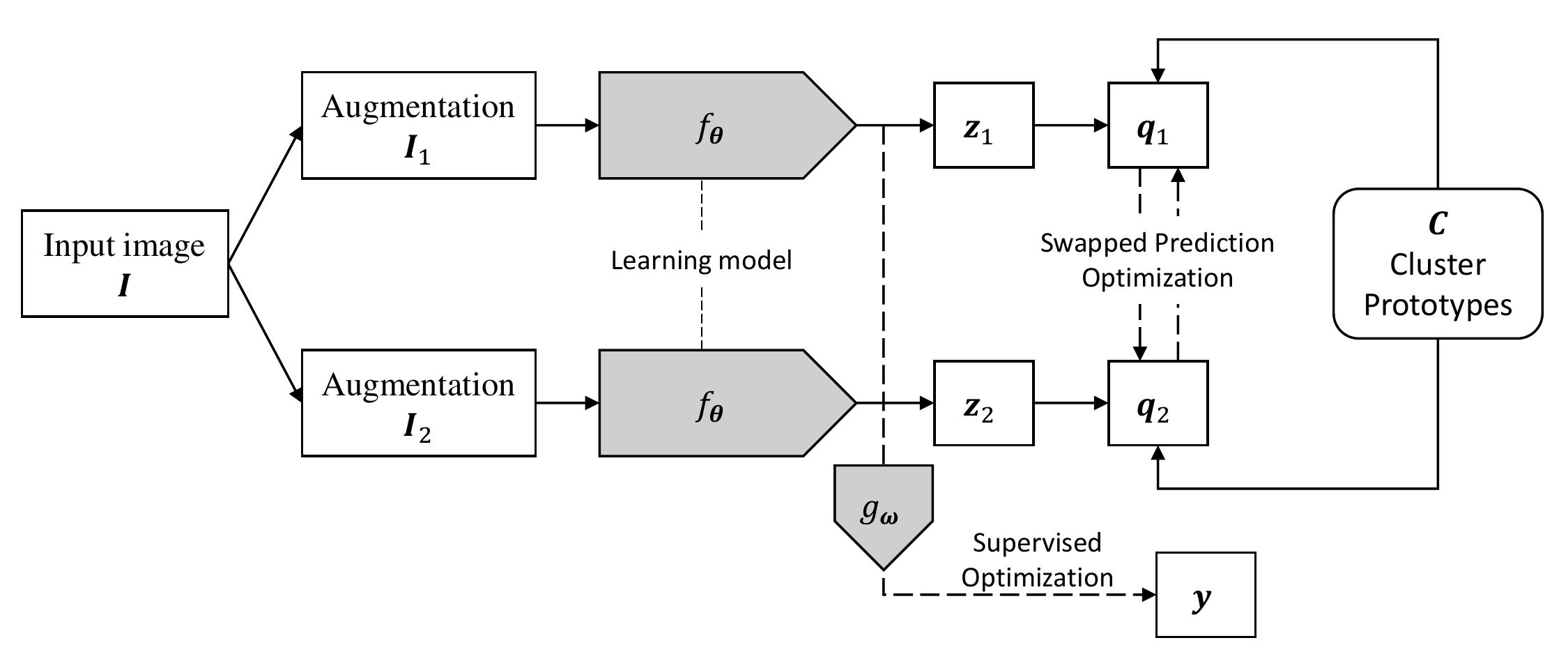}
\centering
\caption{Schematic overview of the training methodology. A given input image $\vec{I}$ is randomly transformed based on the set of augmentation operations $\mathcal{A}$ to $[\vec{I}_1,\vec{I}_2]$. These are processed by the learning model $f_{\vec{\theta}}$ to the features $[\vec{z}_1,\vec{z}_2]$. In turn, these are mapped to their cluster assignments $[\vec{q}_1,\vec{q}_2]$ of the set of $K$ so called prototypes $\{\vec{c}_1,\vec{c}_2,\ldots,\vec{c}_K\}$ and used for optimization in a swapped prediction setting. Existing labels $\vec{y}$ can be leveraged during training.\label{fig:method}}
\end{figure}

We assume that a dataset is given which we denote as $\mathcal{D}=\left[\vec{x}_1,\vec{x}_2,\ldots,\vec{x}_{N'},(\vec{x}_{N'+1},\vec{y}_{N'+1}),\ldots,(\vec{x}_{N},\vec{y}_{N})\right]$, $N$ signal samples, e.g., 2D or 3D images $\vec{x}_k = \vec{I}_k$; or images accompanied by non-imaging information such as text, audio, etc. $\vec{x}_k = \{\vec{I}_k,\vec{\delta}_k\}$; $1 \le k \le N$. A subset of $\mathcal{D}$ consists of $N-N'$ samples which are paired with labels $\vec{y}_k$ ($N' < k \le N$) such as: binary image labels, masks, etc. While the extension to support non-imaging information as input can be realized, e.g., by using robust feature fusion~\cite{Kim2019}, we focus here on learning only from image signal. We propose to use this dataset for hybrid self-supervised / supervised model pretraining, to learn rich, representative features that can be transferred to downstream use-cases, i.e., used as initialization in a supervised training routine. Figure~\ref{fig:method} provides an overview.

\subsection{Online Clustering - Swapped Prediction Optimization}

Following~\cite{Caron2020}, we use an online clustering strategy coupled with principles of contrastive learning to learn image features in a self-supervised way. Given a family of image augmentation operations $\mathcal{A}$ (described later in section~\ref{subsec:augmentation}), the goal is to estimate the visual features parametrized by $\vec{\theta}$ in the model / projector $f_{\vec{\theta}}$ (as shown in Figure~\ref{fig:method}) via assignment to cluster codes. In particular, this assignment is optimized to be invariant to various hierarchies of augmentation operations sampled from $\mathcal{A}$ and applied to any given image $\vec{I}\in\mathcal{D}$. The workflow is as follows:
\begin{enumerate}
    \item Based on an arbitrary image $\vec{I}\in\mathcal{D}$, two transformed images $\left[\vec{\hat{I}}_1,\vec{\hat{I}}_2\right]$ are computed using random augmentation operations sampled from $\mathcal{A}$ and applied hierarchically;
    \item The nonlinear model / projector $f_\theta$ (i.e., the model that we attempt to pretrain) is applied on $\left[\vec{\hat{I}}_1,\vec{\hat{I}}_2\right]$ to compute features $[\vec{z}_1,\vec{z}_2]$ which in turn are assigned to cluster codes $[\vec{q}_1,\vec{q}_2]$ from a set of $K$ prototypes $\{\vec{c}_1,\vec{c}_2,\ldots,\vec{c}_K\}$ ($K$ is a parameter set by the user);
\end{enumerate}

Given the pair of features $[\vec{z}_1,\vec{z}_2]$ and code pair $[\vec{q}_1,\vec{q}_2]$, the self-supervised optimization criterion is formulated using a swapped prediction strategy based on the cross-entropy loss function:
\begin{equation}
\label{eq:lossss}
    \begin{split}
        \mathcal{L}_{ss}(\vec{x}) &= \mathcal{L}(\vec{z}_1,\vec{z}_2)\\
        &= -\sum_i \vec{q}_2^{(i)}\log\vec{p}_1^{(i)} -\sum_i \vec{q}_1^{(i)}\log\vec{p}_2^{(i)}\\
        &= -\sum_i \vec{q}_2^{(i)}\log\frac{\exp\left(\frac{1}{\tau}\vec{z}_1^\top\vec{c}_i\right)}{\sum_j\exp\left(\frac{1}{\tau}\vec{z}_1^\top\vec{c}_j\right)}\\&-\sum_i \vec{q}_1^{(i)}\log\frac{\exp\left(\frac{1}{\tau}\vec{z}_2^\top\vec{c}_i\right)}{\sum_j\exp\left(\frac{1}{\tau}\vec{z}^\top_2\vec{c}_j\right)}
    \end{split}
\end{equation}
where $\tau\in\mathbb{R}$ is a temperature parameter, and $\mathcal{L}_{ss}$ refers to self-supervised loss. Without loss of generality, $\vec{z}_1 = f_{\vec{\theta}}(\vec{\hat{I}}_1) / \|f_{\vec{\theta}}(\vec{\hat{I}}_1)\|_2$ (similar derivation also for $\vec{z}_2$) and all prototype vectors $\{\vec{c}_1,\vec{c}_2,\ldots,\vec{c}_K\}$ are trainable. Following the notation proposed in~\cite{Caron2020}, let~$\vec{C}$ denote a matrix with column vectors defined by the $K$ prototypes. The optimization described in Equation~\ref{eq:lossss} is performed using stochastic batch-wise sampling of cases from the training set $\mathcal{D}$:
\begin{equation}
\label{eq:optss}
    [\vec{\theta}^*,\vec{C}^*] = \argmin_{\vec{\theta},\vec{C}} \frac{1}{N}\sum_{k=1}^{N}\mathcal{L}_{ss}(\vec{x}_k).
\end{equation}

In the following sections we describe in detail the online clustering algorithm, based on two different scenarios: 1) $\mathcal{D}$ consists only of images of one modality (e.g., radiography); and 2) the dataset contains images of multiple modalities (e.g., radiography, ultrasonography, computed tomography, magnetic resonance imaging, etc.).

\subsubsection{Single-Modality Clustering}

Assume a batch-size of $B$ samples is used for training. In the following, we focus on one branch of the processing steps depicted in Figure~\ref{fig:method}; all projectors are shared on the other branch (without loss of generality, let that be $\vec{I}\rightarrow\vec{I}_1\rightarrow\vec{z}_1;\vec{q}_1$). The set of computed output features $\vec{z}$ is captured by matrix $\vec{Z}\in\mathbb{R}^{F\times B}$, where $F$ denotes the size of the any given feature $\vec{z}$ (column vector). The prototypes are captured by matrix $\vec{C}\in\mathbb{R}^{F\times K}$, where $K$ denotes the number of prototypes. Finally, the codes that enable the mapping of projected features to prototypes are captured by matrix $\vec{Q}$ of size $K\times B$. In order to prevent a trivial solution that would map all images in one batch to the same code, an equipartition constraint is enforced based on the entropy measure:
\begin{equation}
\label{eq:Q}
    \max_{\vec{Q}\in\mathcal{Q}}\text{Tr}(\vec{Q}^\top\vec{C}^\top\vec{Z}) + \epsilon H(\vec{Q}),
\end{equation}
where $H$ denotes the entropy and $\epsilon$ the regularization weight~\cite{Caron2020}. Inspired by the work of Asano et al.~\cite{Asano2020}, we follow~\cite{Caron2020} in constraining the solution space $Q$ to ensure that each prototype is selected at least $B/K$ times in one batch. In addition, empirical evidence indicates that using continuous codes is more effective in the online training setting, compared to discretizing the solution. Following the derivation of~\cite{Caron2020} and optimal transport theory, the solution $\vec{Q}^*\in\mathbb{R}^{K\times B}$ to Equation~\ref{eq:Q} can be determined as a normalized exponential matrix using the Sinkhorn-Knopp algorithm~\cite{Cuturi2013}.

\subsubsection{Multi-Modality Clustering}

Training with images from multiple modalities is more challenging. While in theory it may enable the pretraining of robust, modality-invariant rich features that would generalize to a variety of downstream tasks; in practice, simply mixing images of multiple modalities in one single batch impacts the training stability. We hypothesize that a modality specific clustering can alleviate this issue. Let $\mathcal{D}$ contain images of $M$ different modalities. In this case we propose to partition $\mathcal{D} = \bigcup_{m=1}^M\mathcal{D}^{(m)}$, where $\mathcal{D}^{(m)}$ denotes the set of all images on modality indexed by $1 \le m \le M$ (e.g., radiography); and $M$ denotes the number of different modalities captured in $\mathcal{D}$. Without loss of generality, let us assume the batch-size $B$ is a multiple of the number of modalities $M$. We propose to partition each batch of samples in $M$ subsets of equal size, each subset containing only images of one modality sampled randomly from any $\left\{\mathcal{D}^{(1)},\mathcal{D}^{(2)},\ldots,\mathcal{D}^{(M)}\right\}$. In this case, Equation~\ref{eq:Q} can be adapted to:
\begin{equation}
\label{eq:QM}
    \sum_{m=1}^M\max_{\vec{Q}_m\in\mathcal{Q}}\text{Tr}(\vec{Q}_m^\top\vec{C}_m^\top\vec{Z}^{(m)}) + \epsilon H(\vec{Q}_m),
\end{equation}
with $\vec{Q}_m, \vec{C}_m$ conditioned on modality $m$; and $\vec{Z}^{(m)}$ denoting the aggregate of $\frac{B}{M}$ vectors $\vec{z}$ associated with modality $m$ (the remaining variables follow the same definition as in~\ref{eq:Q}). The same logic can be applied to adapt Equation~\ref{eq:lossss}.

\subsection{Hybrid Self-Supervised -- Supervised Learning}

As we defined dataset $\mathcal{D}$, $N-N'$ cases are associated with labels, e.g., provided by human (expert) annotators, extracted via natural language processing, or other automatic means from the image, associated clinical reports, or other corresponding non-imaging data. Recall, for any arbitrary training sample $\vec{x}_k$ $(k > N')$, we denote the corresponding label as $\vec{y}_k$. We propose to dynamically learn from these labels in a joint self-supervised / supervised strategy. Let $\vec{g}_{\vec{\omega}}$ be a deep neural network projector parametrized by $\vec{\omega}$, mapping for any such sample $\vec{x}_k$ from features of model $\vec{f}_{\vec{\theta}}$ (output features and/or intermediate layer features) to an output $\hat{\vec{y}}_k$:
\begin{equation}
    \hat{\vec{y}}_k = \vec{g}_{\vec{\omega}}(\vec{x}_k|\vec{f}_{\vec{\theta}}), \forall k > N'.
\end{equation}
In this case, the goal is similar to any supervised learning problem, i.e., minimize the distance of $\hat{\vec{y}}_k$ to $\vec{y}_k$ according to a loss function $\mathcal{L}_{sup}$ (\emph{sup} $\equiv$ supervised):
\begin{equation}
\label{eq:optsup}
    \left[\vec{\theta}^*,\vec{\omega}^*\right] = \argmin_{\vec{\theta},\vec{\omega}} \frac{1}{N-N'}\sum_{k=N'+1}^{N}\mathcal{L}_{sup}\left[\vec{g}_{\vec{\omega}}(\vec{x}_k|\vec{f}_{\vec{\theta}}), \vec{y}_k\right].
\end{equation}

We combine Equations~\ref{eq:optss} and~\ref{eq:optsup} to a single global optimization criterion, re-balancing the contribution of each using factors $\alpha,\beta\in\mathbb{R}$:
\begin{equation}
\label{eq:main}
\begin{split}
    \left[\vec{\theta}^*,\vec{\omega}^*,\vec{C}^*\right] = &\argmin_{\vec{\theta},\vec{\omega},\vec{C}}\frac{1}{N}\sum_{k=1}^N
    \alpha\mathcal{L}_{ss}(\vec{x_k})\\ 
    &+\mathbf{1}_{k>N'}\beta\mathcal{L}_{sup}\left[\vec{g}_{\vec{\omega}}(\vec{x}_k|\vec{f}_{\vec{\theta}}), \vec{y}_k\right].
\end{split}
\end{equation}

\subsection{Augmentation Strategies}
\label{subsec:augmentation}

In order to determine the input to the model during training $[\vec{I}_1,\vec{I}_2]$, random augmentation operations are applied from a set of augmentation operations defined as $\mathcal{A}$. These operations are as listed below.\smallskip

\textbf{Image rescaling} is applied to a fraction $s$ of the size of the original image $\vec{I}$ as $\Psi_s(\vec{I})$, with $\Psi$ denoting the rescaling function and $s\in\mathbb{R}$ sampled uniformly at random from the interval $[0.1,0.7]$. The rescaling factor $s$ is sampled for large range of possible values to encourage the learning of robust, scale-invariant features.\smallskip

\textbf{Energy-based augmentation} is performed based on the image normalization algorithm proposed by Philipsen et al.~\cite{Philipsen2015}. Following their methodology, an image $\vec{I}$ is decomposed into $B$ energy bands $\vec{I}^{(1)},\vec{I}^{(2)},\ldots,\vec{I}^{(B)}$ using Gaussian filtering. For each band $1 \le i \le B$, the energy value $e_i(\vec{I},\Omega)$ is computed as the brightness dispersion in a predefined image region defined by $\Omega$ (in our case the entire image). Following~\cite{Philipsen2015}, the normalized image is computed as:
\begin{equation}
    \vec{\hat{I}}(\Omega) = \sum_{i=1}^{B}\lambda_i(\Omega)\vec{I}^{(i)}=\sum_{i=1}^{B}\frac{e_i^{ref}(\Omega)}{e_i(\vec{I},\Omega)}\vec{I}^{(i)},
\end{equation}
where $e_i^{ref}(\Omega)=\frac{1}{R}\sum_{k=1}^{R}e_i(\vec{I}_k,\Omega)$ denotes the reference energy value on band $i$, with $\vec{I}_1,\vec{I}_2\ldots\vec{I}_R$ denoting $R$ pre-selected reference images. We set $R=1000$ and propose to augment the image using a variable reference energy $e_i^{ref}(\Omega)$ for any given band $1 \le i \le B$ around the mean value. Concretely, on each band we propose to model the distribution of the $R$ reference values using a Gaussian distribution and sample the value of the reference energy from the range $[-\sigma,+\sigma]$ around the mean energy value.\smallskip

\textbf{Intensity rescaling} is applied in two different ways: 1) nonlinear rescaling using a gamma transform with the exponent $\gamma\in\mathbb{R}$ sampled uniformly at random from the interval $[1.8,2.6]$; and 2) linear rescaling of the intensity as $a*\vec{I} + b$ with $a\in[0.9,1.1]$ a random uniform sample and $b$ restricted to $\pm20\%$ of the intensity range of $\vec{I}$.\smallskip

\textbf{Cropping} from random image locations (sampled uniformly at random) is the final augmentation applied.

\section{Experiments and Results}
\label{sec:experiments}

\subsection{Datasets for Self-Supervised Training}

We constructed several datasets for self-supervised training (based on 2D and 3D image modalities). The weights of models trained on these datasets using self-supervision were then transferred to initialize models that were optimized using supervision on downstream tasks. The datasets are the following:

\begin{itemize}
    \item[--] 2D X-ray dataset $\mathcal{D}_{X}$ containing 1,297,699 X-ray images capturing various anatomies, including chest, spine, back, arm, leg, and more. The data is acquired from both public~\cite{MIMIC1data,MIMIC2data,TBdata,INDIANAdata,BIMCVdata,LIDCdata,SHENZENMONTdata,BONEAGEdata,MURAdata,NIHdata} and internal sources.
    \item[--] 2D mixed modality dataset $\mathcal{D}_{M}$ containing 105,006,320 images/slices of various anatomies (head, abdomen, chest, legs, etc.) and from various imaging modalities, including the x-ray dataset $\mathcal{D}_{X}$. Except the public data contained in $\mathcal{D}_{X}$ as described in the previous section, this dataset contains only internal data. The proportions per modality are: 72\% computed tomography (CT) slices, 25\% magnetic resonance imaging (MRI) slices, around 1\% X-ray and the rest ultrasonography (US) images.
    \item[--] 3D computed tomography (CT) dataset $\mathcal{D}_{CT}$ containing 24,440 3D CT volumes coming from 1,345,040 DICOM slices of non-contrast CT head scans aquired from internal sources.
\end{itemize}

\subsection{Training Hyper-Parameters, Infrastructure and Scaling}
Different architectures have been investigated as part of our experiments, in the 2D context all variants of residual networks~\cite{He2016} (including ResNet-152 and ResNet-50 with several variants denoted as ResNet-50w2/w4 as described in~\cite{Caron2020}). Training hyper-parameters are defined in Table~\ref{table:hypers}. Several parameters are inherited from SwaV~\cite{Caron2020}, for model details we refer the reader to that reference. The training infrastructure consists of 4 nodes (each with 8 Volta GPUs with 16GB GPU memory, 80 cores and 512 GB main memory). All nodes are connected via InfiniBand. The system uses the Quobyte file system for parallel and distributed IO operation and PyTorch distributed functionality is applied to scale the training to multiple nodes.

\begin{table}[t]
\centering
\caption{Hyper-parameters for self-supervised training in 2D/3D.\label{table:hypers}}
\begin{tabular}{L{3cm} L{2cm} L{2.5cm}}
Parameter&2D-Experiments&3D-Experiments\\
\midrule
Number of crops&[2,6]&[1,4]\\
Size of crops&[224,92]&[(224,192),(112,96)]\\
Image size range&[600,1200]&[(112,96),(250,200)]\\
Cropped image border&[0.1,0.1]&[0.1,0.1]\\
Assigned crops&[0,1]&[0,1]\\
Temperature&0.1&0.1\\
Epsilon&0.03&0.03\\
\# Sinkhorn iterations&3&3\\
Feature dimension&128&128\\
\# Prototypes&3000&1500\\
Queue length&3840&1920\\
Epoch queue starts&10&3\\
Epochs&100&100\\
Batch size&32&4\\
Start Learning Rate&0.6&0.6\\
Final Learning Rate&0.0006&0.0006\\
Freeze prototypes&10000&4000\\
Weight Decay&1e-06&1e-06\\
\bottomrule
\end{tabular}
\end{table}

\subsection{Abnormality Detection in Chest Radiography}

\begin{figure}[t]
\includegraphics[width=8.8cm]{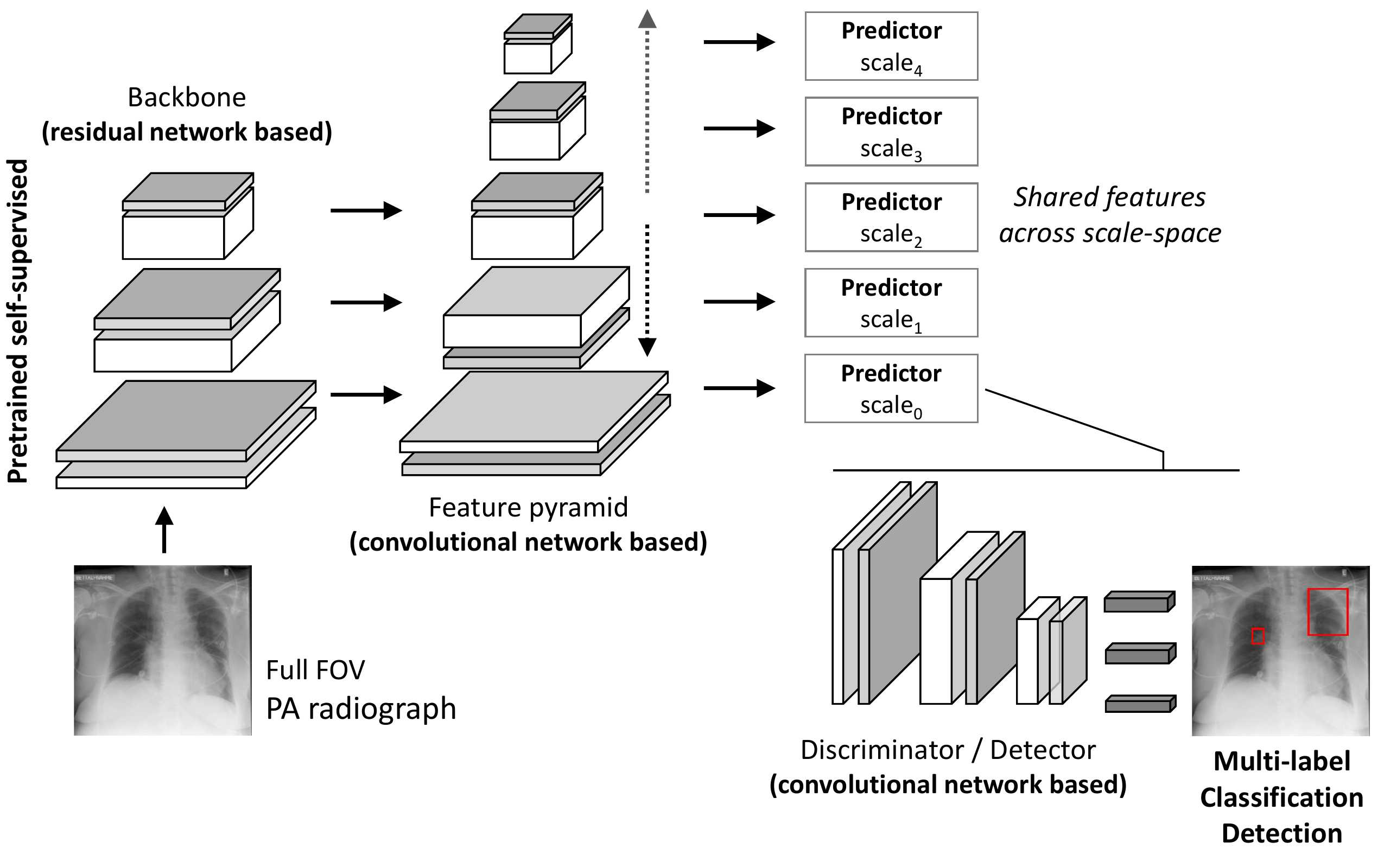}
\centering
\caption{Architecture used for classification and detection of lesion and pneumothorax in chest radiographs. The backbone (ResNet-50w2) is pretrained using self-supervision.\label{fig:xraymodel}}
\end{figure}

\begin{figure}[t]
\includegraphics[width=8.5cm]{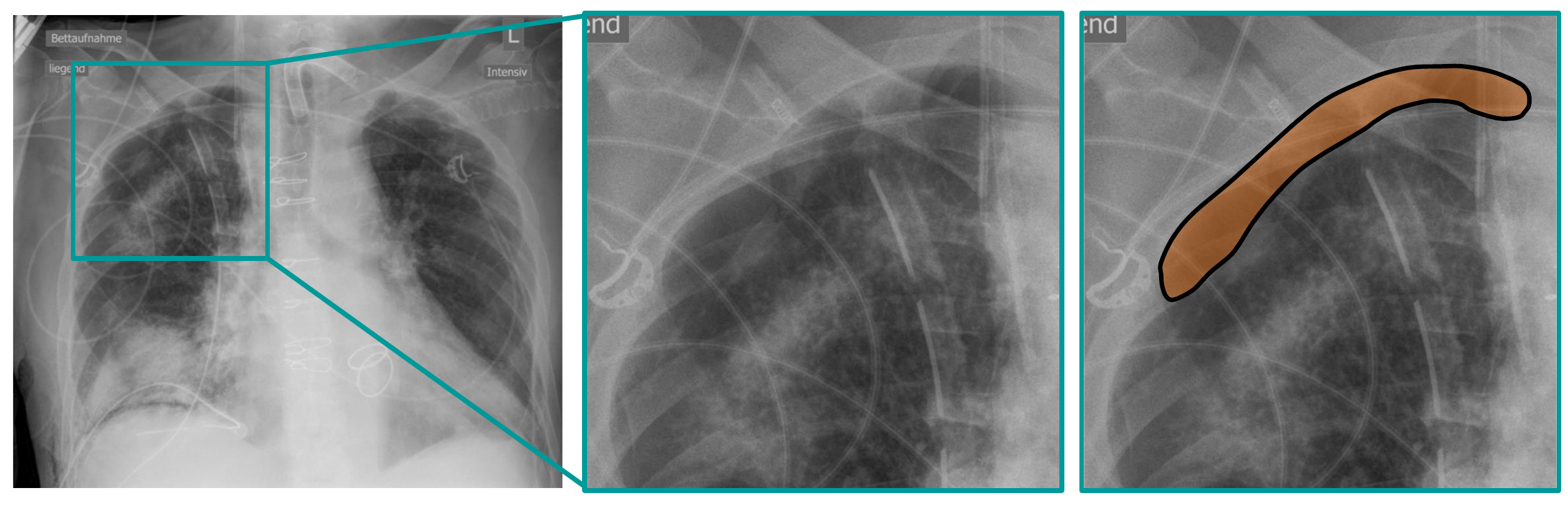}
\centering
\caption{Example case from the pneumothorax test set. \textbf{Left}: Chest radiograph with right upper zone showing a lucent area towards the apex with vaguely defined pleural line suggestive of pneumothorax. ICD tube along with numerous lines make the distinction of the pneumothorax difficult; \textbf{Middle}: Image sub-region capturing the pneumothorax; \textbf{Right}: Curve highlighting the dehiscent visceral pleural separated from the thoracic wall - indicating the pneumothorax.\label{fig:pneumothorax}}
\end{figure}

\begin{figure}[t]
\centering
\includegraphics[width=9cm]{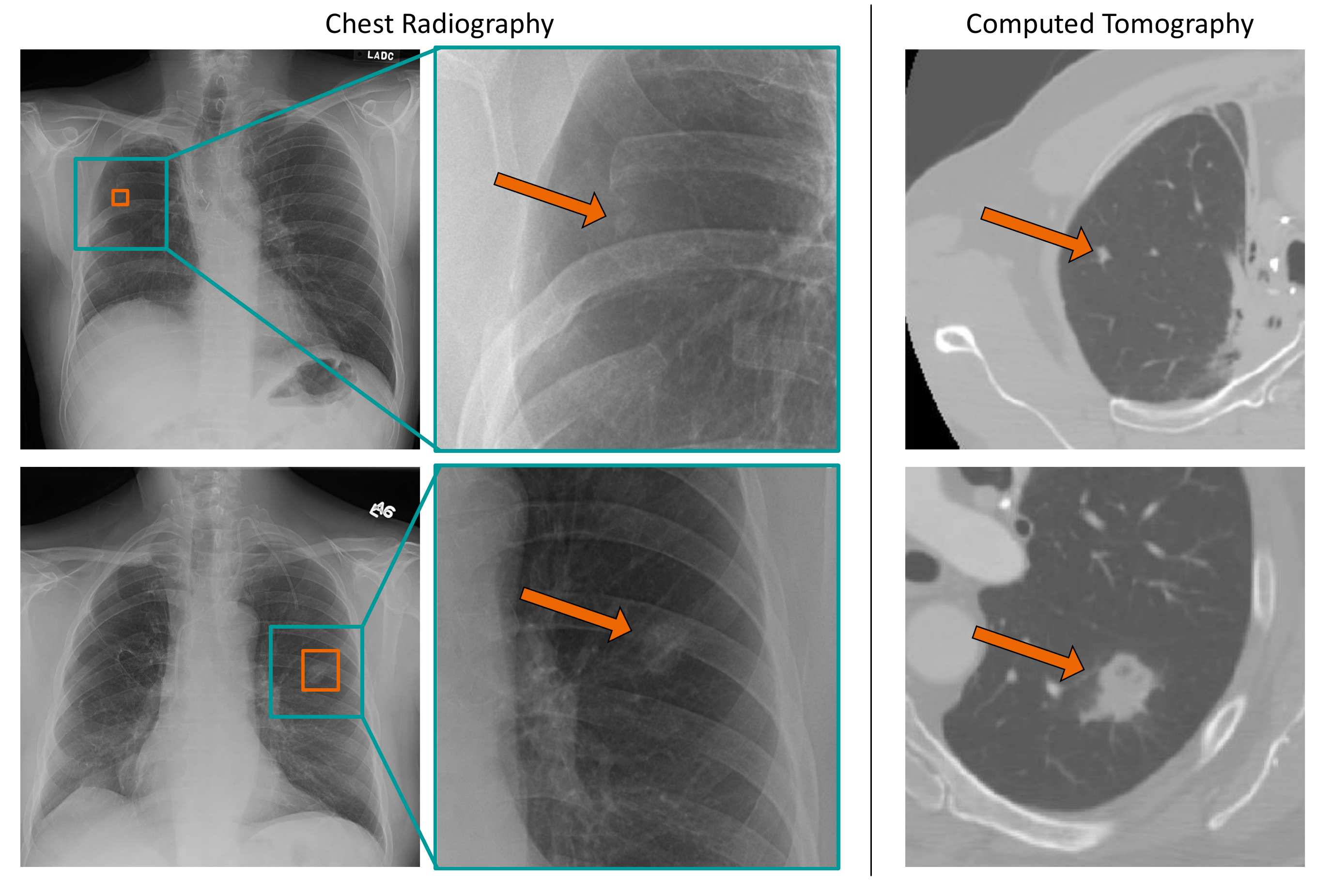}
\caption{Example cases from lesion test set. \textbf{Left}: Chest radiographs with lesions highlighted using a red bounding box; \textbf{Middle}: Image sub-regions with arrows indicating the lesions. \textbf{Right}: Axial slices of the corresponding CT scan with arrows indicating the same lesions. The first row shows an image of a patient who has sustained a right thoracotomy. In the resection area there is a very subtle lesion that was marked positive only in \textit{LIDC-synch}. In the generation of \textit{LIDC-staged} it was missed by all three readers when reading the radiograph to generate candidates for CT-confirmation, and as such the case is marked as negative in \emph{LIDC-staged}. In contrast, the lesion captured in the second row is much larger and easier to see. It is marked as positive in both \textit{LIDC-synch} and \textit{LIDC-staged}.\label{fig:lesion}}
\end{figure}

We focused on the detection of lung lesions (i.e., nodules/masses) and pneumothorax from frontal chest radiographs. These are critical findings, lesions with potential long-term relevance (e.g., pulmonary malignancy, cancer) while pneumothorax as acute finding often immediately endangers the life of the patient. As proposed in our previous work~\cite{Ghesu2021EU,Ghesu2021PTX}, we approach the problem as a multi-class detection problem using bounding boxes to isolate the abnormalities. The model architecture is fully-convolutional and multi-scale, and is inspired by~\cite{Tian2020}. The entire content of the image is processed in one single forward pass and labeled bounding boxes (with an associated probability) around relevant abnormalities are predicted (see Figure~\ref{fig:xraymodel}). A total of 11730 chest radiographs were used for training. The data was acquired from internal data sources. Various abnormalities (including lesion and pneumothorax) were annotated by expert radiologists on each image using bounding boxes (see Figures~\ref{fig:pneumothorax} and~\ref{fig:lesion}). Further details related to the training data and training routine can be found in our previous work~\cite{Ghesu2021EU,Ghesu2021PTX}.

For testing the pneumothorax detection feature we use a test set of 321 chest radiographs, all acquired at bedside in anterior-posterior view -- the majority capturing severely ill patients, covered by tubes and/or wires with 34 cases acquired in the ICU. The ground-truth is determined by a consensus of 3 expert radiologists. Each image is first read independently by each reader, followed by a joint discussion to determine the final annotation. As pneumothoraces can be very small (to a few millimeters in sectional width) or obscured by other structures, e.g., the ribs, they can easily be overlooked. By using three readers, we intend to minimize that risk. Of the 321 radiographs, 125 are identified as pneumothorax positive using 148 bounding boxes to isolate the abnormal anatomy. Figure~\ref{fig:pneumothorax} shows an example.

For testing the lesion detection feature we use a test set of 288 radiographs from the LIDC dataset~\cite{LIDCdata}. Each radiograph is paired with a CT scan acquired in close time-proximity. The information from this additional modality is used to improve the ground-truth quality. As not all lesions captured in CT are also visible in chest radiography~\cite{Barbosa2021}, we propose two protocols to derive two versions of the ground-truth:
\begin{enumerate}
    \item Synchronous reading of chest radiograph and corresponding CT by an expert radiologist, marking on the radiograph only lesions that are visible. We denote this version of the ground-truth as \textit{LIDC-synch}.
    \item A staged approach is applied. In the first stage, 3 expert radiologists read independently all chest radiographs and mark lesions (without looking at CT). Subsequently all marks are aggregated by an additional radiologist, removing any duplicate marks of the same lesion. These candidate marks are then assessed in a second stage using CT as point of reference - for each mark on the radiograph, if the CT displays a lesion at that location that mark is positive; if not, the mark is removed. We denote this version of the ground-truth as \textit{LIDC-staged}.
\end{enumerate}

With 146 (out of 288) positive radiographs and 187 lesions marked, \emph{LIDC-synch} is a significantly more sensitive ground-truth than \emph{LIDC-staged} (111 positive radiographs with 133 lesions marked). However, \emph{LIDC-staged} has the benefit of removing any bias related to the assessment of visibility on radiographs (a step performed in the derivation of \emph{LIDC-synch}). Example images and annotations are shown in Figure~\ref{fig:lesion}.

ROC-AUC is used to assess the classification performance, the accuracy of the bounding box detection is assessed using fROC. In particular we report the sensitivity at instance level (captured in the $y$-axis in fROC) averaged at the following average numbers of false-positive per image (captured in the $x$-axis in fROC): $[0.01, 0.05, 0.125, 0.25, 0.5, 1.0, 2.0, 4.0, 8.0]$. We compare our solution with 4 alternative approaches:
\begin{enumerate}
    \item Supervised learning~\cite{He2016} on the ImageNet dataset~\cite{ImageNet};
    \item Self-supervised learning~\cite{Caron2020} on the ImageNet dataset (denoted as SwaV);
    \item SimCLR method for self-supervised learning~\cite{Chen2020} on the internal dataset;
    \item Using no pretraining, relying on a random initialization of the network weights (in our experience, often used for medical image analysis applications).
\end{enumerate}

For more simplicity in the interpretation of the results, we often focus the comparison of our method with method 1) proposed by He et al.~\cite{He2016} -- the previously best strategy for this task. Table~\ref{table:performancexray} gives an overview of the performance of all reference methods.

\begin{table}[t]
\centering
\caption{AUC performance for lesion detection on \emph{LIDC-staged} when using 100\%, 50\%, 25\% or 10\% of the training data. Selection of the subsets is done randomly.\label{table:performancexray}}
\begin{tabular}{C{2.8cm} L{.8cm} L{.8cm} L{.8cm} L{.8cm}}
\multirow{2}{*}{Method} & \multicolumn{4}{c}{AUC Performance (\emph{LIDC-staged})$^1$}\\
\cmidrule{2-5}
& 100\% & 50\% & 25\% & 10\%\\
\midrule
No pretraining & 0.77 & 0.73 & 0.65 & 0.53\\
SimCLR~\cite{Chen2020} & 0.90 & 0.88 & 0.82 & 0.79\\
SwAV~\cite{Caron2020} & 0.90$^2$ & 0.89 & 0.85 & 0.80\\
Supervised NI~\cite{He2016} & 0.91$^2$ & 0.89 & 0.82 & 0.80\\
\textbf{Ours} & \textbf{0.94}$^2$ & \textbf{0.91} & \textbf{0.85} & \textbf{0.85}\\
\midrule
\multicolumn{5}{l}{$^1$Average of 5 models; selected with highest AUC on validation set}\\
\multicolumn{5}{l}{$^2$Over 5 training rounds standard deviation of AUC $\le$0.003 for each,}\\
\multicolumn{5}{l}{indicating the high stability of the performance measurement}\\
\end{tabular}
\end{table}

\begin{figure*}[t]
\centering
\includegraphics[width=5cm]{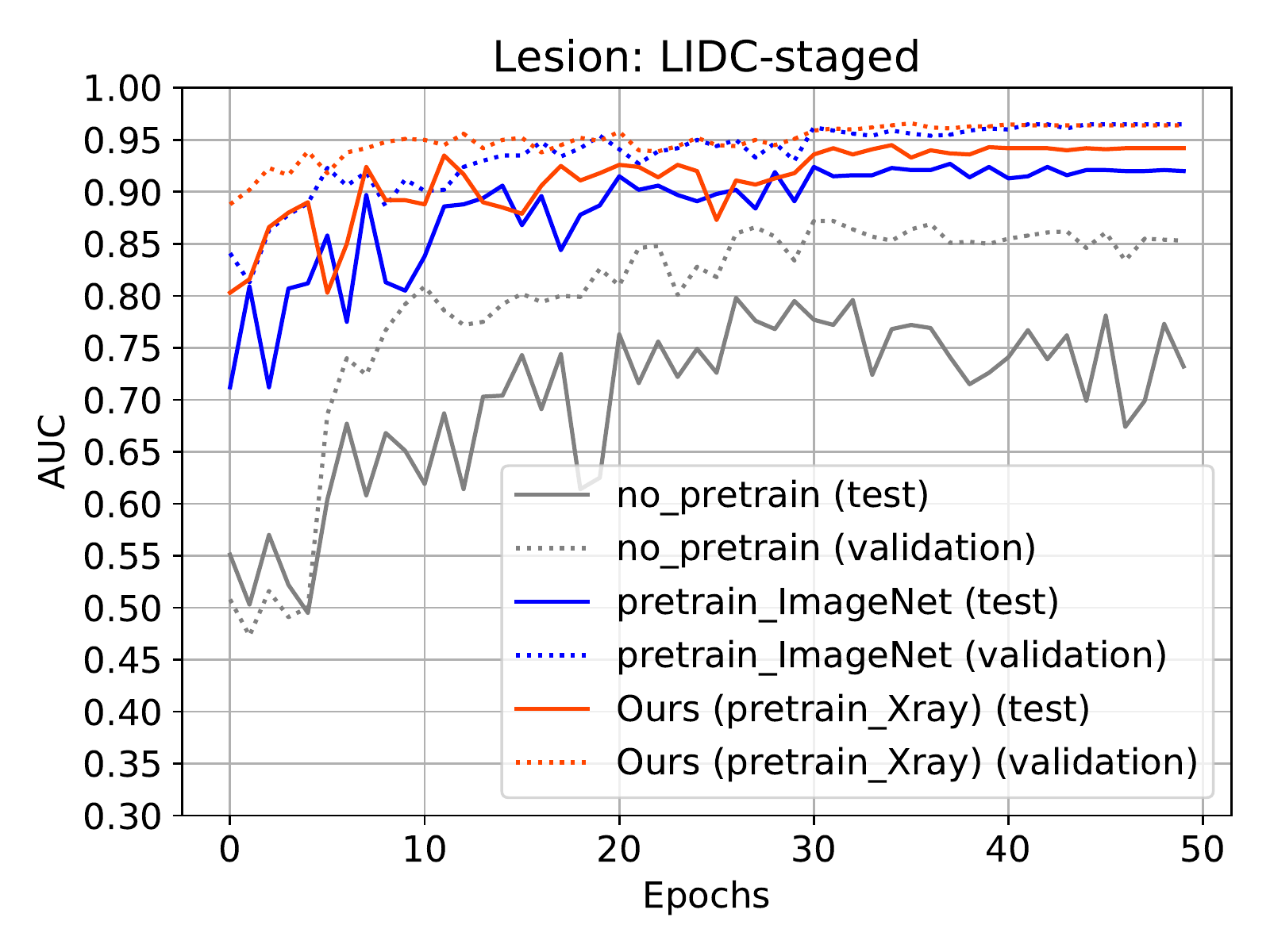}
\includegraphics[width=5cm]{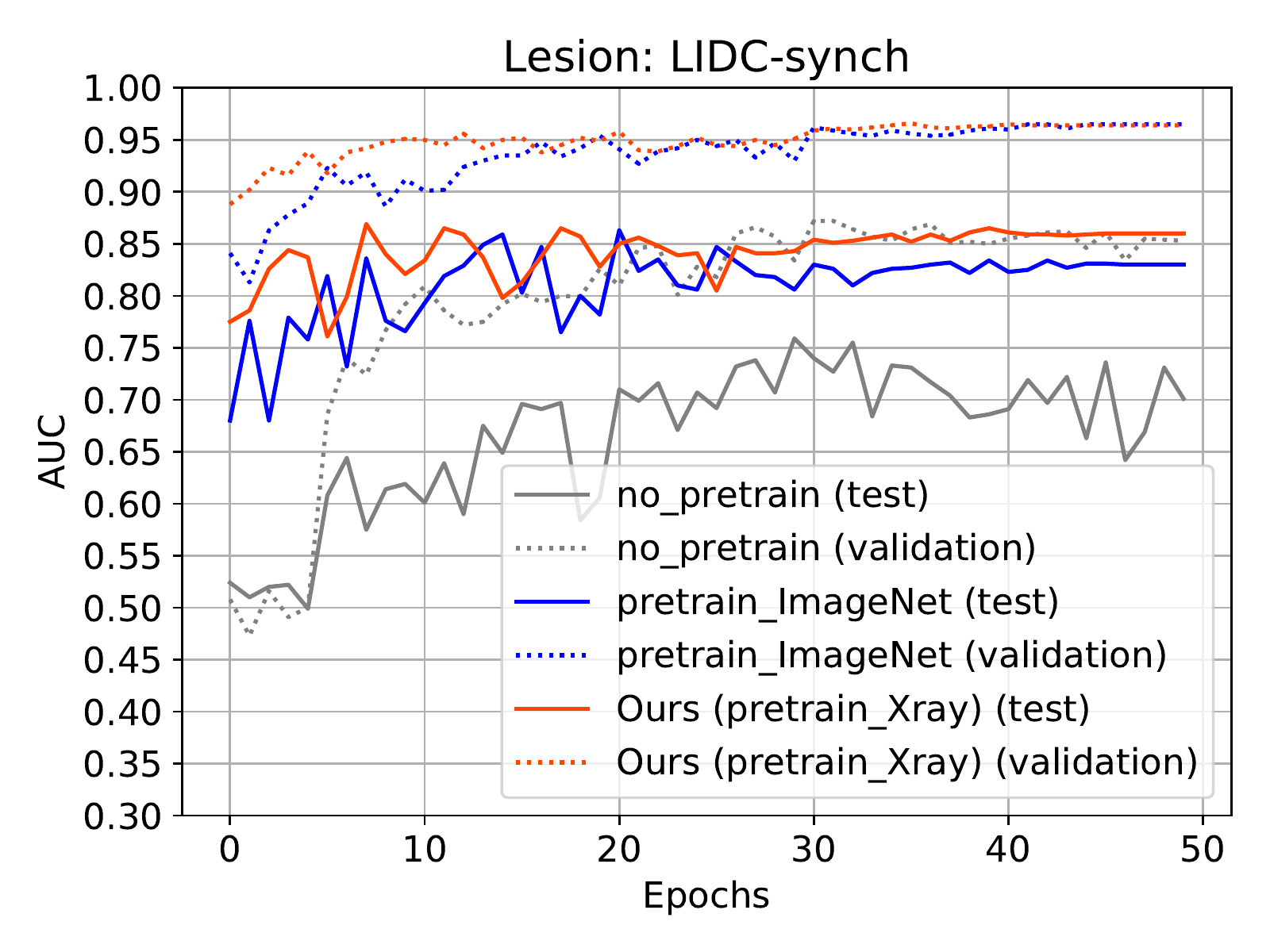}
\includegraphics[width=5cm]{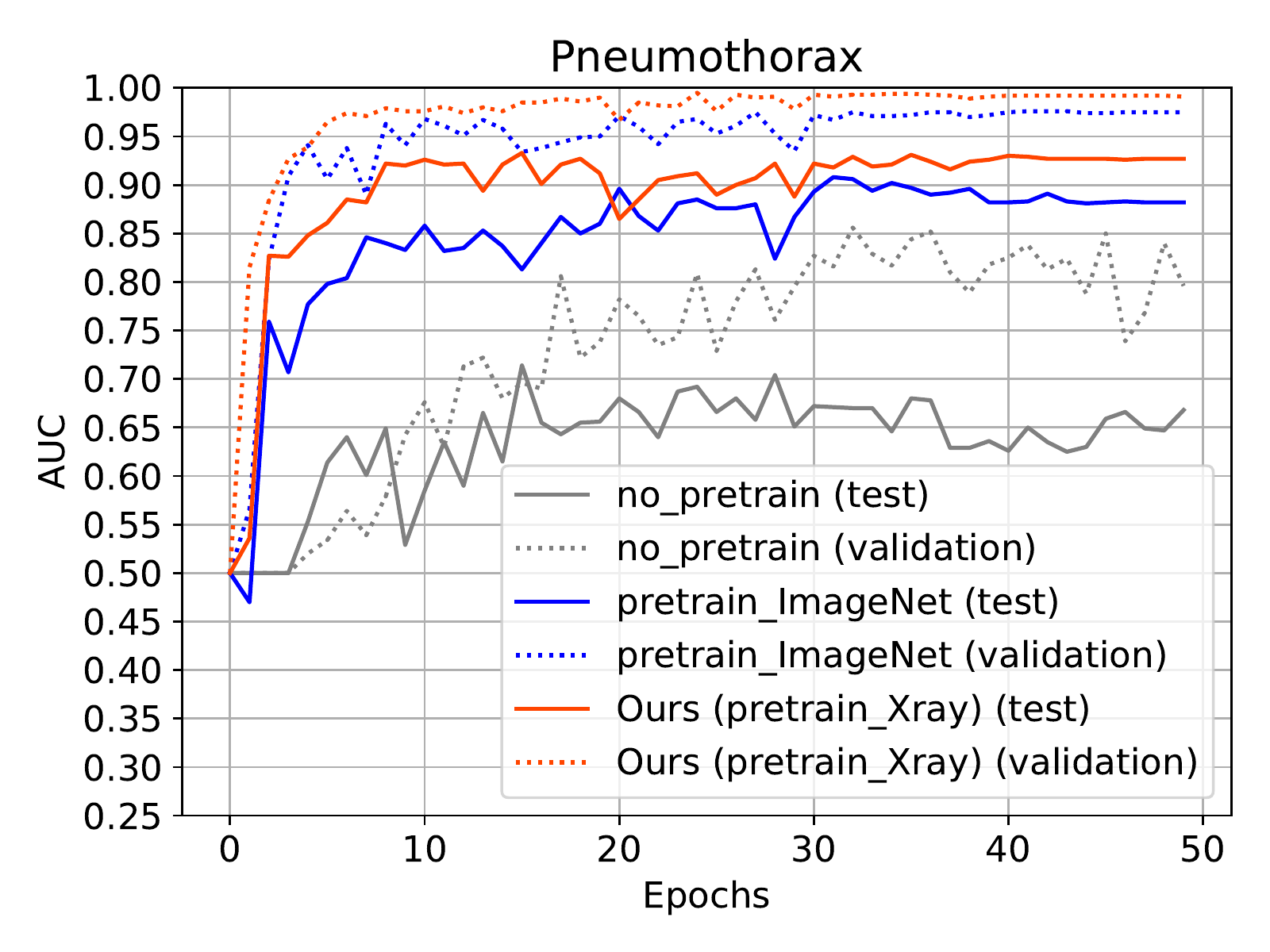}
\includegraphics[width=5cm]{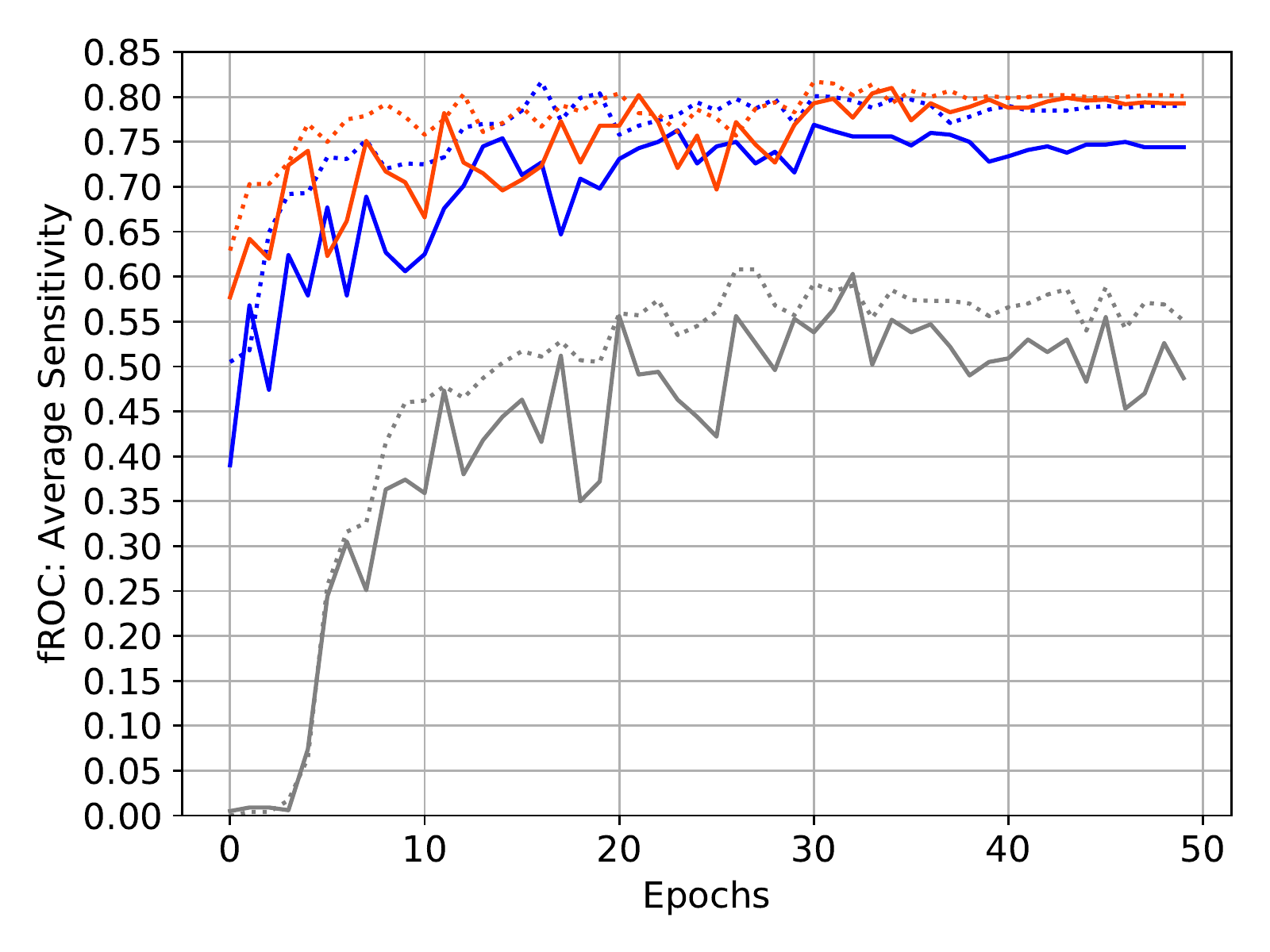}
\includegraphics[width=5cm]{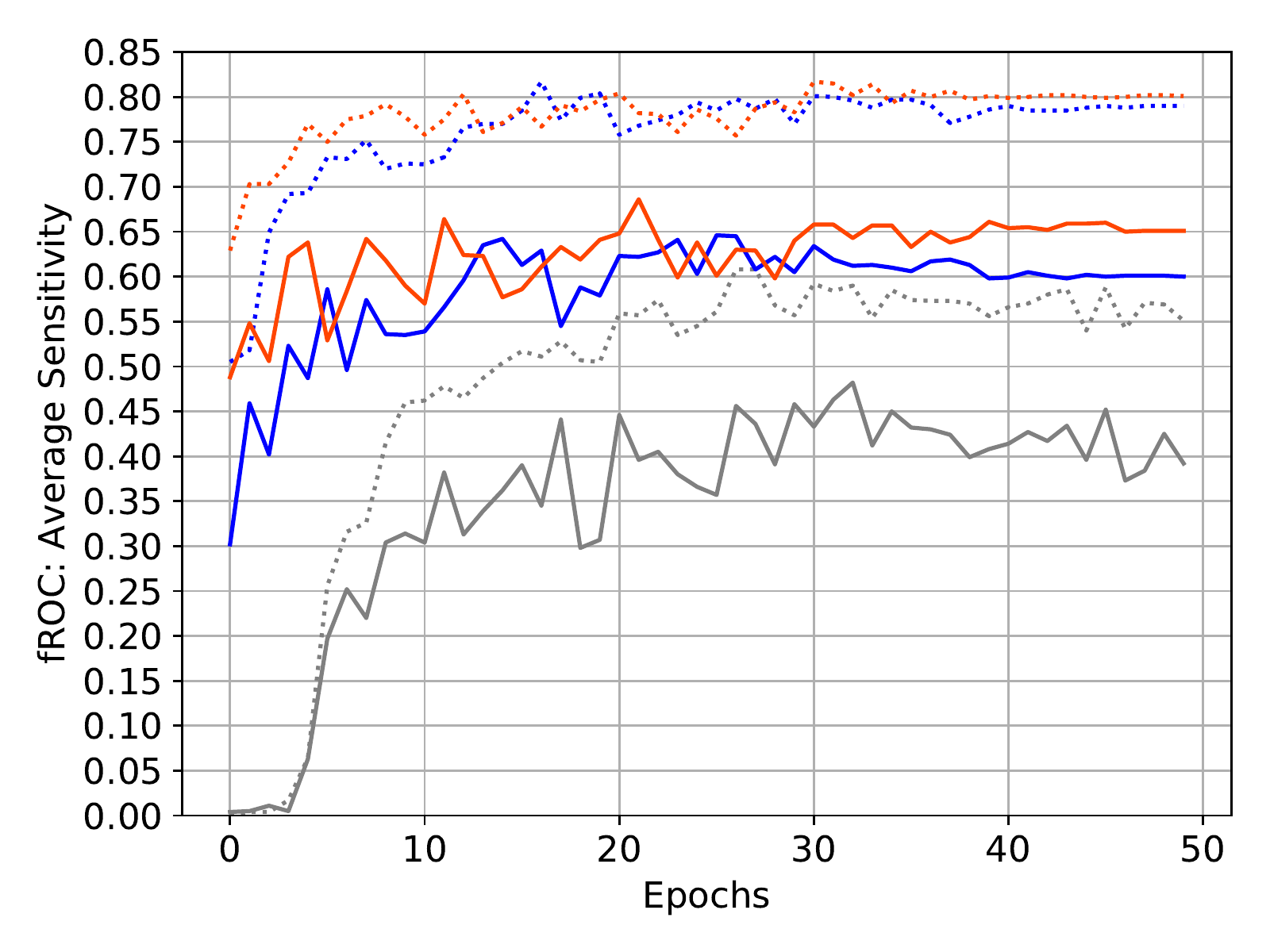}
\includegraphics[width=5cm]{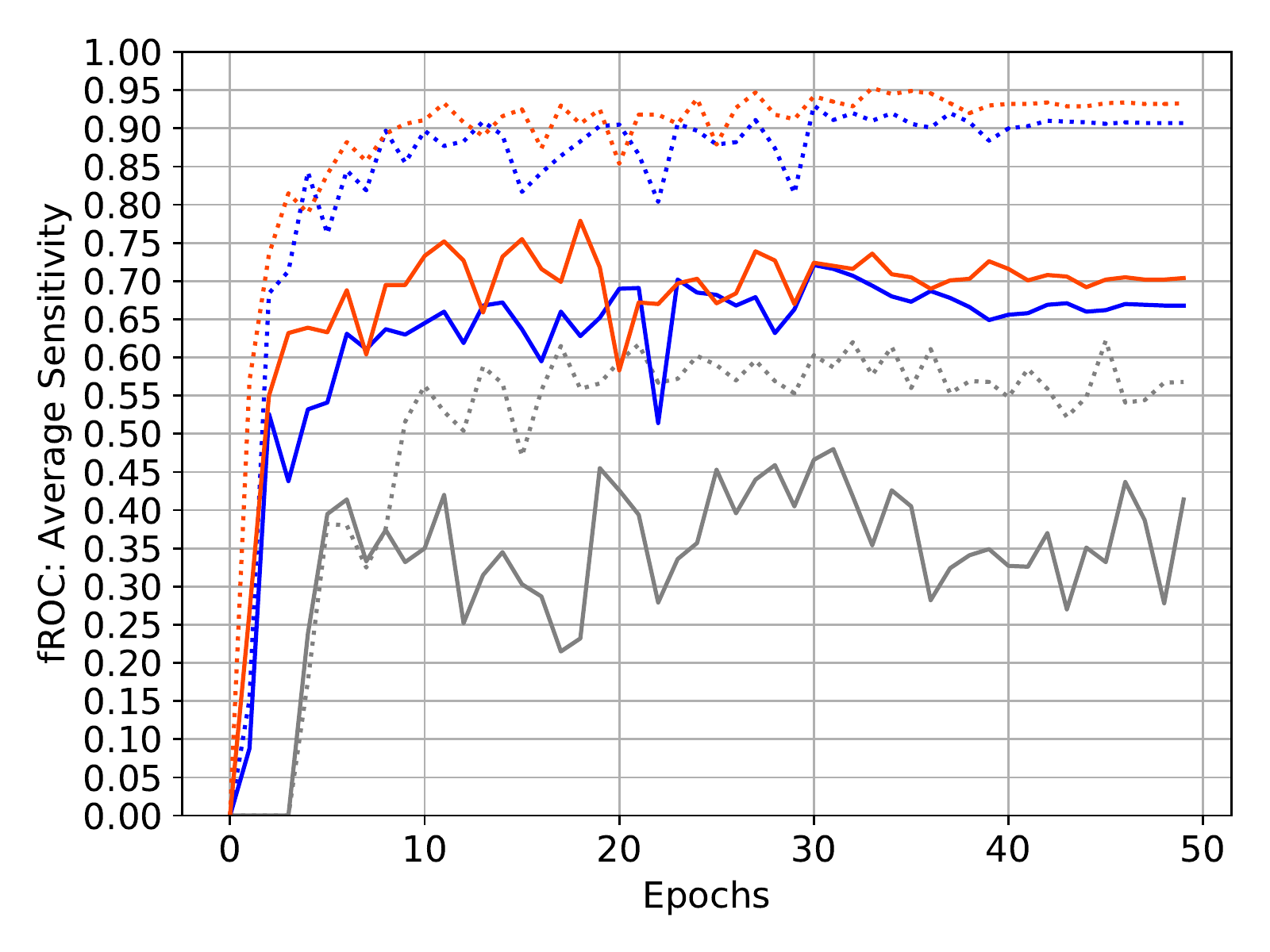}
\caption{Performance evolution of lesion and pneumothorax detection in chest radiographs during training (\textbf{left}: lesion on \emph{LIDC-staged}, \textbf{middle}: lesion on \emph{LIDC-synch} and \textbf{right}: pneumothorax). Our solution (initializing the model with self-supervised pretrained weights on the X-ray dataset $\mathcal{D}_{X}$) significantly outperforms both in terms of AUC and average instance detection sensitivity the previously best pretraining strategy, i.e., supervised pretraining on ImageNet~\cite{He2016}. The difference is significant, ranging between 3 - 5\%. The difference is much larger when compared to using no-pretraining - ranging between 20 - 25\% on the lesion test and almost 30\% on the pneumothorax test data.\label{fig:xrayperformance}}
\end{figure*}

Figure~\ref{fig:xrayperformance} shows the performance evolution for lesion and pneumothorax detection on the test and validation sets during training. We highlight the AUC and average instance level fROC sensitivity as a function of the epochs. A significant increase in average performance is achieved for the test set.

\begin{figure}[t]
\includegraphics[width=6cm]{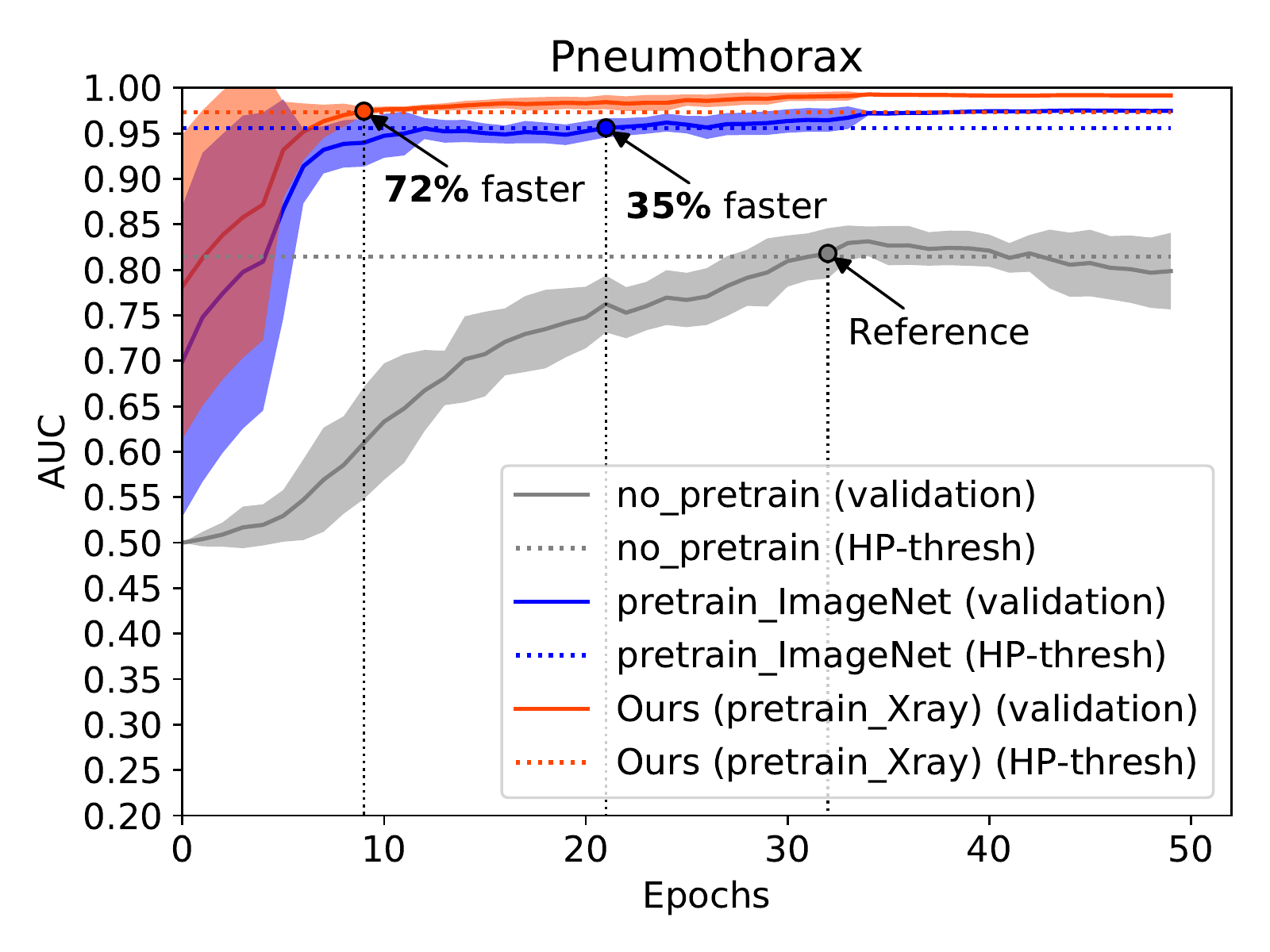}
\centering
\caption{Visualization of the training speed-up in model for pneumothorax. We denote the convergence point as the earliest epoch during training where 98\% of the final best performance is achieved (denoted as HP-thresh, i.e., high performance threshold). Using our method, one can achieve convergence 72\% faster than using no pretraining. The transparent area along each curve denotes the standard deviation.\label{fig:xrayspeed}}
\end{figure}

\begin{figure}[t]
\includegraphics[height=3cm]{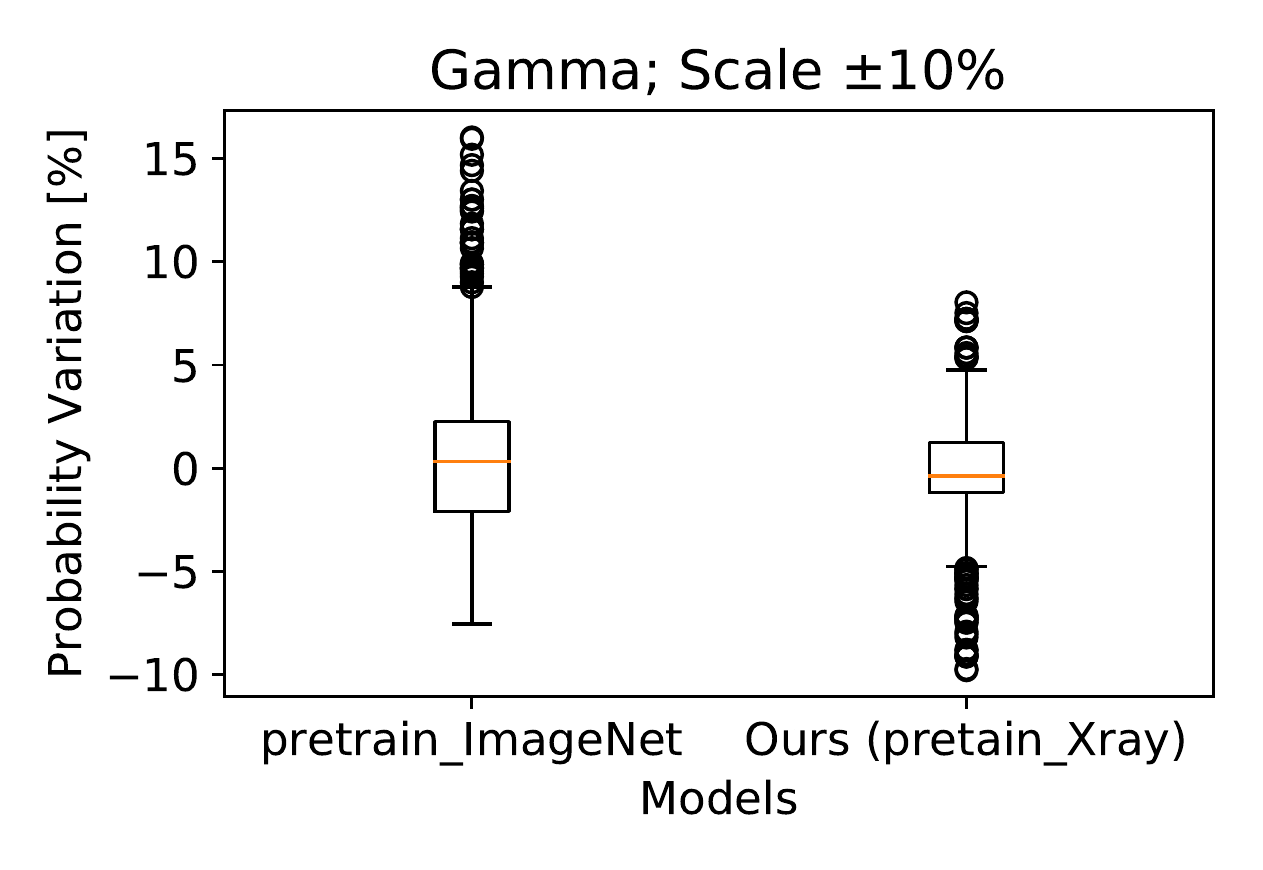}\hspace{-0.5em}
\includegraphics[height=3cm]{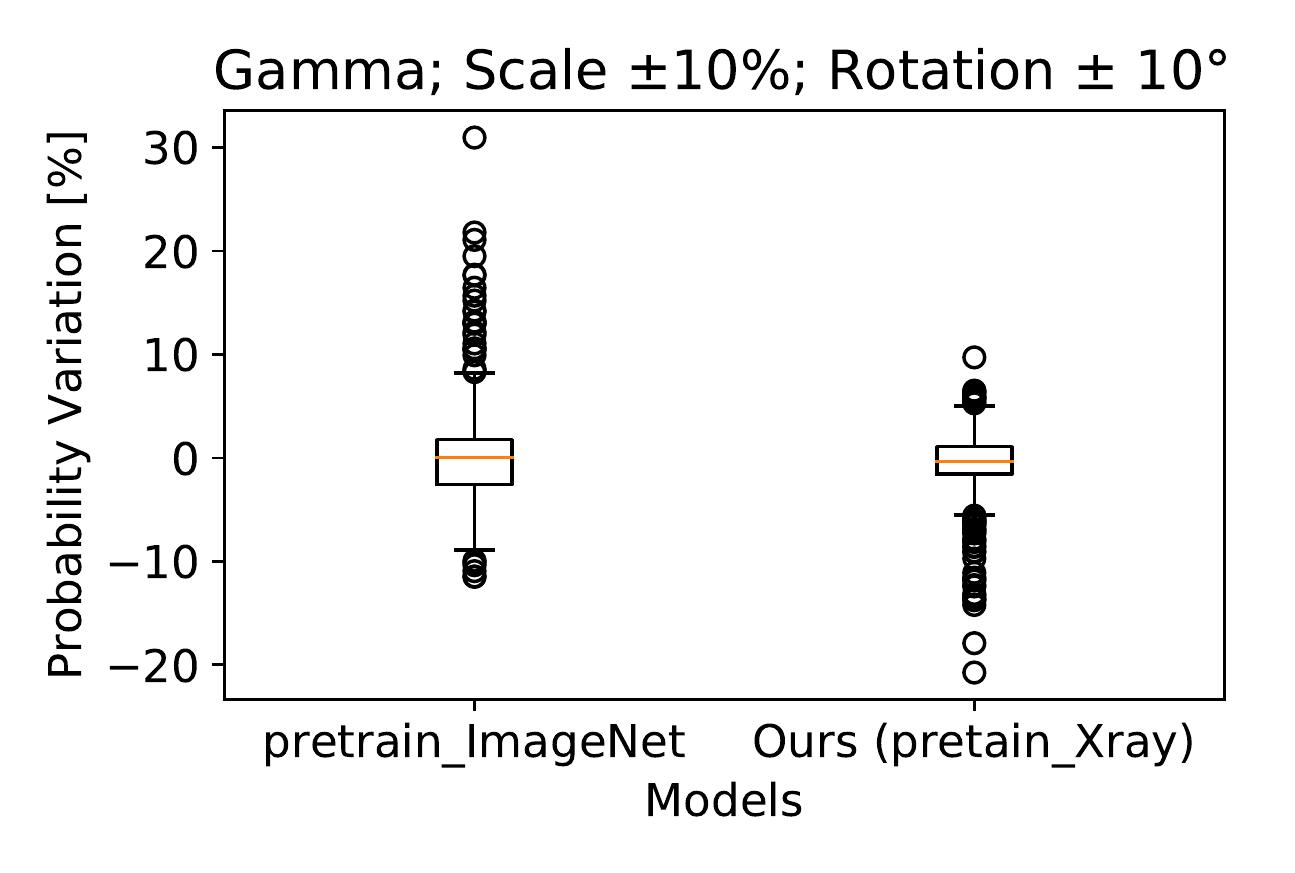}
\centering
\caption{Using self-supervised pretraining on $\mathcal{D}_{X}$ leads in an increase in robustness. The box plot shows the relative deviation in probability for lesion (at case level) when applying various augmentations, such as gamma transform with $\gamma\in[1.8,2.6]$, and/or random image rotation/scaling. The distribution is shown for a random set of 50 radiographs with 50 random transformations per image (i.e., 2500 data points). The reference method refers to~\cite{He2016}.\label{fig:xrayrobustness}}
\end{figure}

Figure~\ref{fig:xrayspeed} shows the acceleration in convergence speed when using our method compared to conventional supervised pretraining or no pretraining. The speedup is at least 50\% (also on both lesion test sets); both when analyzing evolution of performance on test data and validation data. Finally, Figure~\ref{fig:xrayrobustness} highlights the increased robustness of self-supervised pretrained models after down-stream fine-tuning with respect to certain image variations which are typically observed in practice (e.g., image rotations/scaling or intensity variations).

\subsection{Brain Metastases Detection in MRI}
 Automated detection and segmentation of brain metastases in 3D MRI scans could support therapy workflows. In this study, we conducted another experiment on slice-wise brain metastasis on contrast-enhanced magnetization-prepared rapid acquisition with gradient echo (MPRAGE) scans, which can be used for treatment selection, planning and longitudinal monitoring by guiding radiosurgery protocols and other treatment decisions. However, this task remains challenging due, in part, to the scarcity of training data containing metastatic tissue in an MRI volume, which makes learning clinically meaningful image features from scratch challenging. A reliable pretrained model may have the potential to mitigate this limited data problem. Thus we focused on analyzing the impact of self-supervised training as pretraining on classifying 2D slices with metastases in MPRAGE volumes. We utilized a 2.5D encoder-decoder network to first obtain a segmentation mask showing potential areas of suspected metastases~\cite{ghesu2021quantifying}. The output segmentation mask was subsequently used as an attention channel along with the 5 input slices to train a 2.5D classification network to perform a slice-wise classification. The architecture of the classification network followed the concept of ResNet-50w2. The six channel input was compressed to make three channels by a $1\times1$ convolution. Our dataset included a training set (341 cases), a validation set (36 cases) and a test set (43 cases). The details about the data and preprocessing can be found in~\cite{yoo2021evaluating}. 

\begin{figure}[t]
\includegraphics[width=6cm]{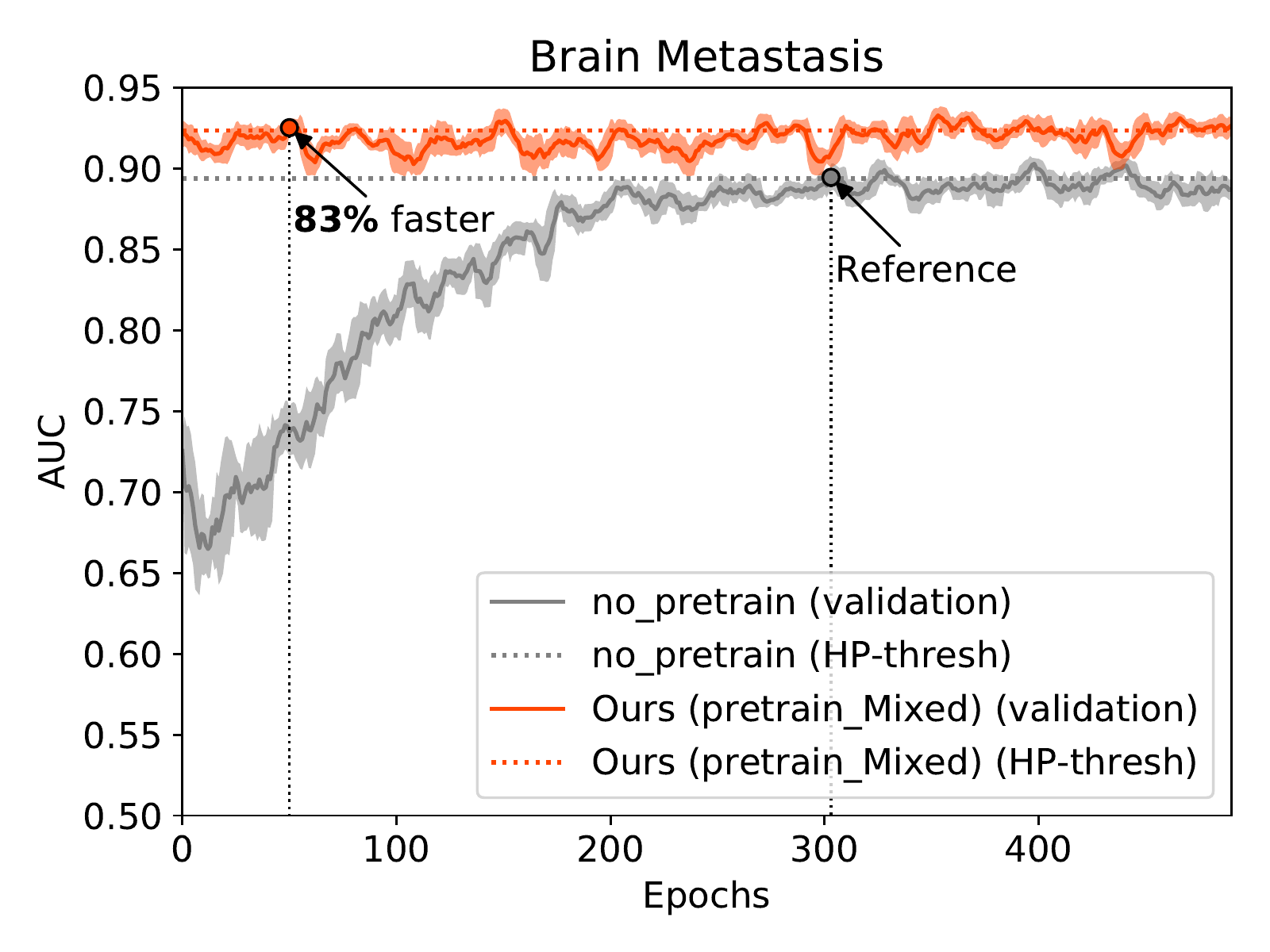}
\centering
\caption{Validation AUC evolution over training epochs for brain metastatic slice detection. Using our self-supervised pretraining leads to an increase in training convergence rate and validation AUC.}
\label{fig:brainmet_trn_speed}
\end{figure}

\begin{figure}[t]
\includegraphics[width=7cm]{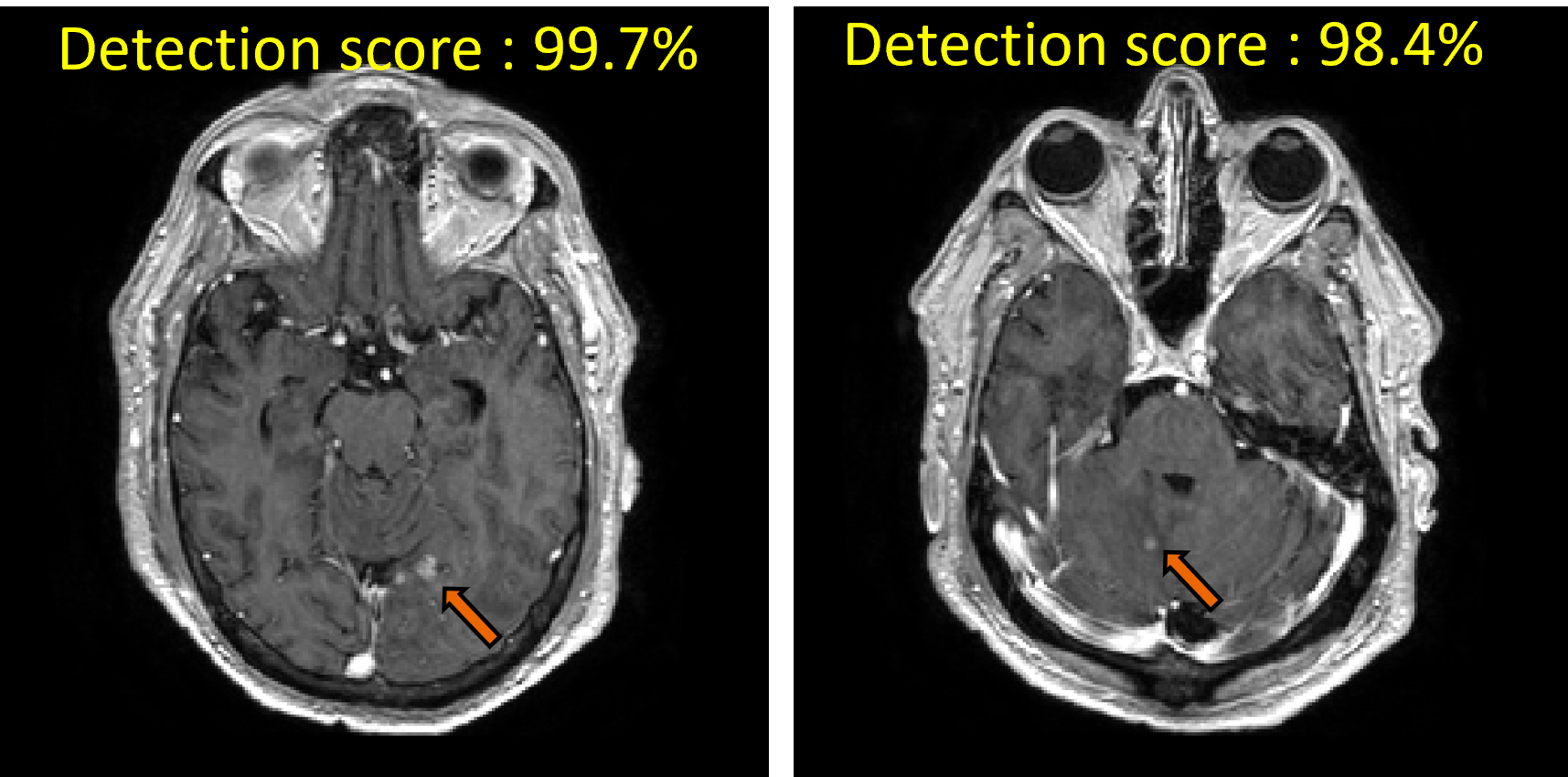}
\centering
\caption{Example post-contrast MPRAGE slices from a patient with brain metastases. The images show metastatic slices and their detection scores by the model pretrained with our self-supervised learning. Image courtesy: University of Michigan.}
\label{fig:brainmet_example}
\end{figure}

The metastatic slice detection performance was evaluated by ROC-AUC and mean average precision. Figure~\ref{fig:brainmet_trn_speed} shows the training evolution using detection AUC measured on the validation dataset for each epoch. Training the model from scratch reached AUC 90\% after 300 epochs. On the other hand, training the model initialized with the pretrained ResNet-50 achieved AUC 92-93\% within 10 epochs. Also the pretrained model consistently outperformed the model without pretraining by AUC 2-3\%. The testing AUC with our pretraining method was 93.2\% which was higher than both the model without pretraining and the model pretrained with SwAV~\cite{Caron2020} by 2\% as shown in Table~\ref{tb:brainmet_accuracy}. Mean average precision with our pretraining method was 85.6\% which was higher than the model without pretraining by 5.6\% and than the model pretrained with SwAV~\cite{Caron2020} by 1.2\%. Both the pretrained models produced accuracy 94.3\% that is higher than the model without pretraining by 3.3\%. Figure~\ref{fig:brainmet_example} shows examples of metastatic slice detection in a patient with brain metastasis.
 
\begin{table}[t]
\centering
\caption{Performance comparison for slice-wise brain metastasis detection on 8145 testing MPRAGE slice images (1336 metastatic images). AUC: area under the ROC-curve, mAP: mean average precision. Accuracy was computed with an operation point 0.6.}
\begin{tabular}{C{2cm} C{1.5cm} C{1.5cm} C{1.5cm}}
\midrule
Method & AUC & mAP & Accuracy \\
\midrule
No pretraining & 0.913 & 0.799 & 0.910 \\
SwAV~\cite{Caron2020} & 0.913 & 0.844 & 0.943 \\
\textbf{Ours} & \textbf{0.932} & \textbf{0.856} & \textbf{0.943} \\
\bottomrule
\end{tabular}
\label{tb:brainmet_accuracy}
\end{table}

\subsection{Brain Hemorrhage Detection in CT}
Current standard of care for evaluation of patients presenting with stroke symptoms or following head trauma involves assessment of non-contrast CT (NCCT) scans for presence of hemorrhage. Accurate and timely detection of head bleeding is critical to start the appropriate treatment as soon as possible such as starting administration of thrombolytics for stroke patients or surgical intervention for trauma patients. Automated detection of hemorrhage in NCCT scans~\cite{Arbabshirani2018, Kuo2019}, has the potential to minimize the time it takes for a patient to receive the appropriate treatment.

\begin{figure}[t]
\centering
\includegraphics[height=3.3cm]{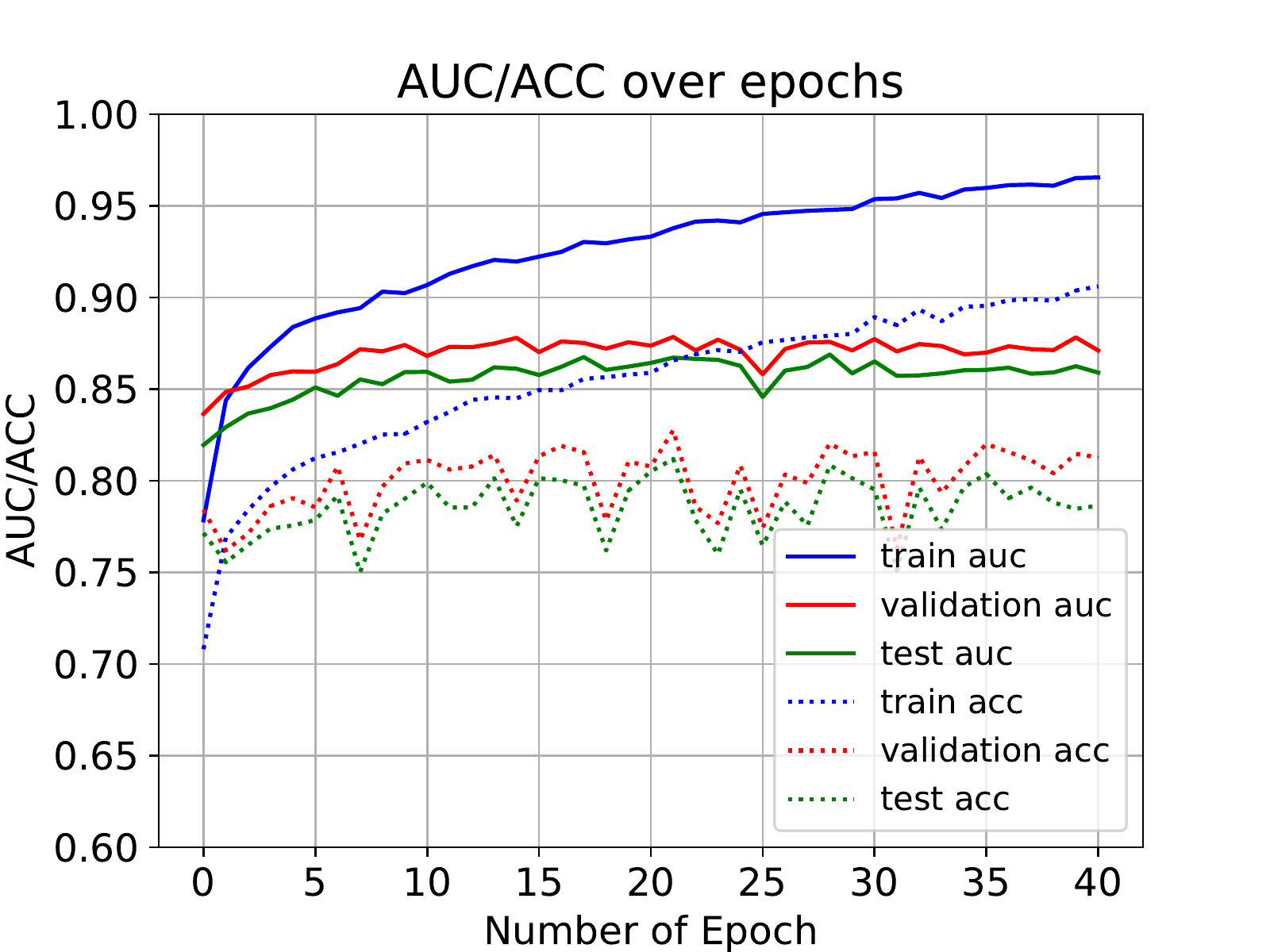}\hspace{-0.5em}
\includegraphics[height=3.3cm]{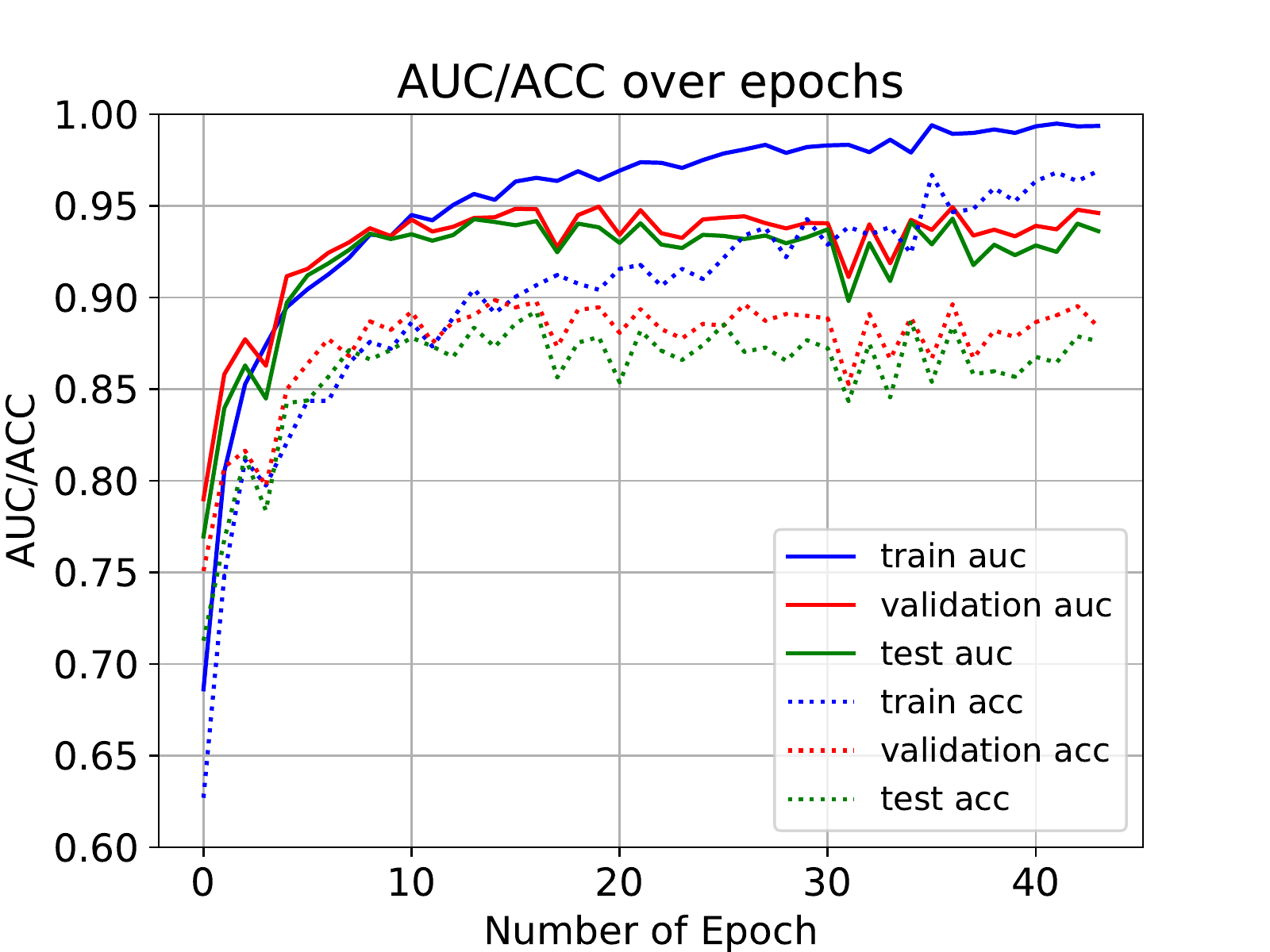}
\caption{Performance of hemorrhage detection on 3D NCCT: Using auxiliary tasks and self-supervised pretraining of a 3D network on $\mathcal{D}_{CT}$ leads in an increase in performance. The auxiliary tasks include the use of hemorrhage types and hemorrhage labels on each slice. The left figure illustrates AUC and accuracy for the base model trained only on 3D hemorrhage labels for training, validation and testing data splits. The right figure illustrates the performance for the model trained using auxiliary tasks and self-supervision. It shows a significant increase of the AUC from 0.88/0.87 (validation/test) to 0.95/0.94.\label{fig:ncct3d}}
\end{figure}

\begin{figure}[t]
\centering
\includegraphics[width=7.0cm]{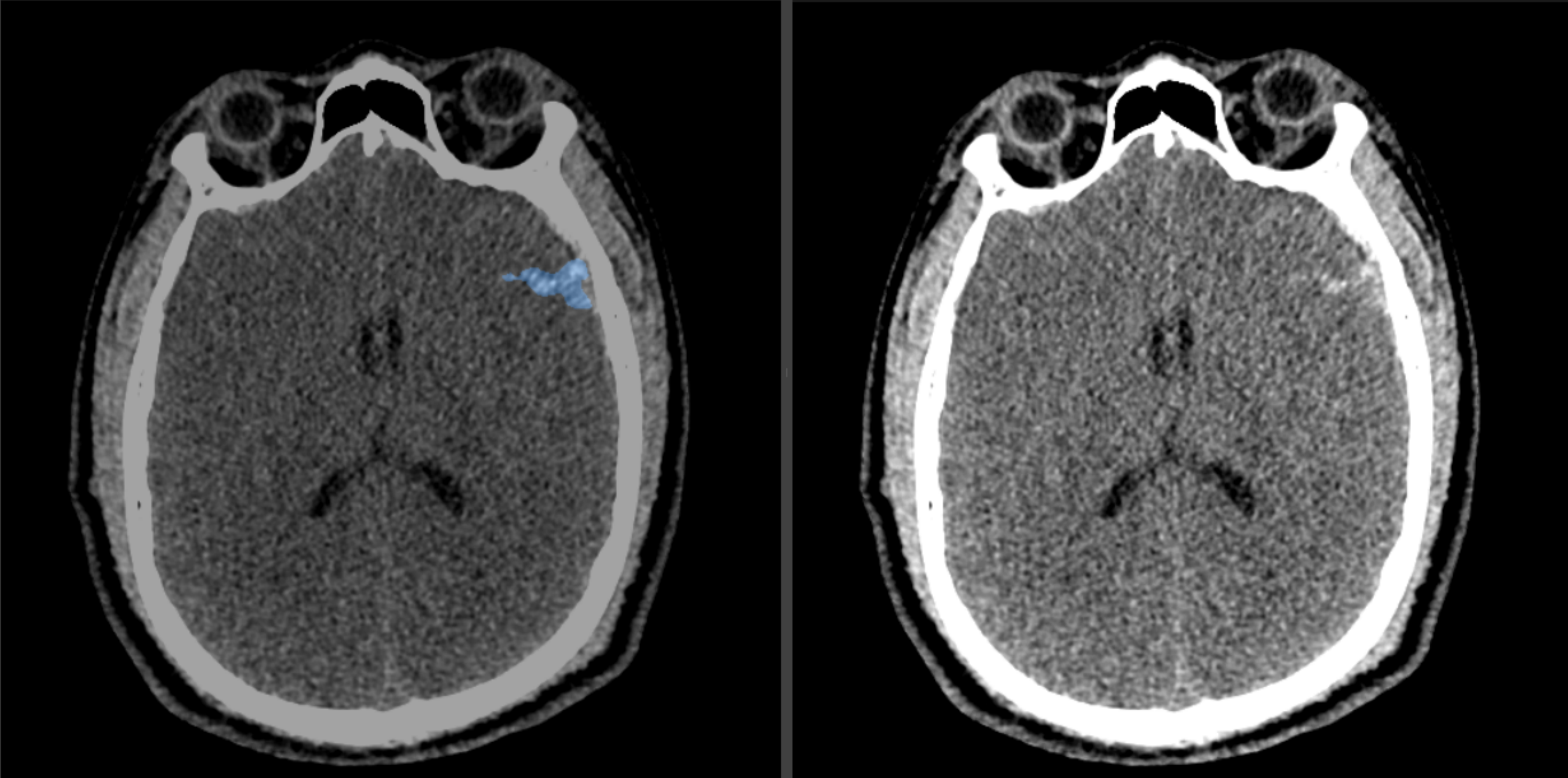}
\includegraphics[width=7.0cm]{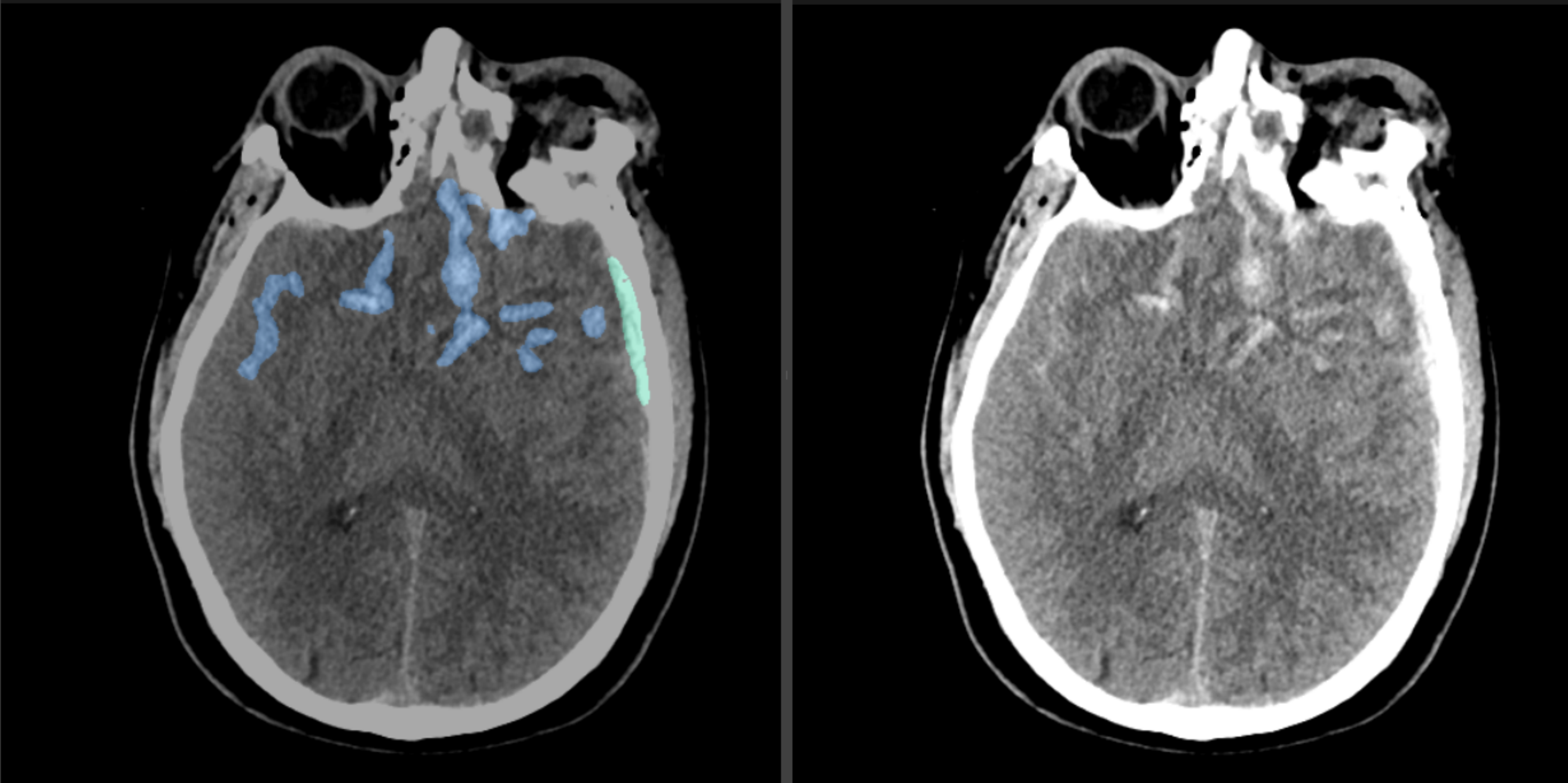}
\caption{Examples of hemorrhage detection on 3D NCCT on various types of hemorrhage. On the left the it is illustrated with color overlay the hemorrhage region and on the right the NCCT image slice.\label{fig:ncct3d_examples}}
\end{figure}

In this work we investigate the gains of automated AI-based detection of hemorrhage in 3D NCCT scans by using both auxiliary tasks as well as self-supervised pretraining of a 3D network. The input dicoms are stacked together in a 3D volume, we then reformat the input volume to be in axial orientation and a set of head landmarks are used to crop the brain region and scale it in a box of dimensions 40x224x192. The Hounsfield units are normalized to (0,1) by using a transformation with a (center, window) = (55, 200). For feature extraction we employ a 3D densely connected network with variable input size and 5 dense blocks with (1,2,3,3,3) units, each unit having 3D Convolution, 3D BatchNormalization and LeakyReLU activation layers with 16 initial features and a growth rate of 7. The first two blocks only process data in-plane with (1,3,3) kernels and downsample only in (x,y) plane and the last three process data with full (3,3,3) 3D kernels. The final features of fixed dimension 1024 are computed through adaptive pooling (both max and average pooling).
A set of 3017 3D volumes are used for validation and model selection and a set of 2945 volumes are used for testing (both datasets coming from patients not included in pretraining). The main task is hemorrhage detection for each 3D volume and we have used as auxiliary tasks training with hemorrage types labels (subarachnoid, subdural, epidural, intraventricular, intraparenchymal hemorrhages) and with presence/absence of hemorrhage for each slice. Figure~\ref{fig:ncct3d} illustrates the AUC and accuracy of the base network and the improvement by training with auxiliary tasks as well as with self-supervised pretraining. The network trained with only presence/absence of hemorrhage for the whole 3D volume achieves an AUC of 0.88/0.87 for the validation/test sets while the network trained also with the auxiliary tasks and the self-supervision achives and AUC of 0.95/0.94 respectively, that is an 8\% increase in performance. Figure~\ref{fig:ncct3d_examples} illustrates different types of hemorrhage successfully detected by the system. Table~\ref{tb:ncct3dabl} shows the performance gains by sequentially using each of the auxiliary tasks and self-supervision -- the best performance is obtained when all are used with self-supervision.


\begin{table}[t]
\centering
\caption{Ablation performance of hemorrhage detection on 3D NCCT showing the AUC for validation/test when only 3D labels are used and adding types, slice labels and the self-supervision pretraining of the same 3D network.}
\begin{tabular}{C{1cm} C{1.2cm} C{1.2cm} C{1.2cm} C{1.4cm}}
\midrule
Labels & 3D & 3D, types & 3D, types, slice & 3D, types, slice, self-supervision \\
\midrule
AUC & 0.87/0.88 & 0.89/0.88 & 0.94/0.93 & 0.95/0.94 \\
\bottomrule
\end{tabular}
\label{tb:ncct3dabl}
\end{table}

\subsection{Directions of Future Research}

Further optimization and research is required in different directions: 1) With a training time of 6.5 - 14 days (depending on the training dataset - $\mathcal{D}_X$, $\mathcal{D}_M$ or $\mathcal{D}_{CT}$) further optimization and better scalability of the training is required to execute more training rounds, and perform more ablative analysis. This has limited the amount of experiments and analysis; 2) Once the previous point is addressed, more work is needed to investigate the effectiveness of the proposed training technique on more diverse models; and 3) More dedicated focus is needed to investigate the utility of self-supervised learning in tracking and registration tasks, in which often models are very small and shallow to ensure high efficiency.

\section{Conclusion}
\label{sec:conclusion}
In conclusion, we propose an effective technique for self-supervised learning based on contrastive learning and online clustering, with support for hybrid self-supervised / supervised learning and multi-modality training data (2D and 3D). We demonstrate the scalability of the method on a large dataset of over 105,000,000 images, highlighting the impact of the learned image representations in improving the accuracy (average of 6-8\% AUC), robustness and training speed (up to 85\%) on various downstream tasks.\smallskip

\noindent\textbf{Disclaimer}\\ 
The concepts and information presented in this paper are based on research results that are not commercially available.
\noindent\textbf{Acknowledgements}\\
The authors acknowledge the National Cancer Institute and the Foundation for the National Institutes of Health, and their critical role in the creation of the free publicly available LIDC/IDRI Database used in this study.
\bibliography{paper}

\end{document}